\documentclass{article}
\PassOptionsToPackage{table}{xcolor}
\usepackage{iclr2027_conference,times}
\usepackage[T1]{fontenc}
\usepackage{amsmath,amssymb,mathtools}
\usepackage{graphicx,booktabs,tabularx}
\usepackage{longtable}
\usepackage{flafter}
\usepackage{xcolor}
\usepackage{tikz}
\usetikzlibrary{arrows.meta,positioning,calc,fit,backgrounds}
\usepackage{algorithm}
\usepackage{algpseudocode}
\usepackage[pdfversion=1.7]{hyperref}
\hypersetup{hidelinks}
\usepackage{url}

\iclrfinalcopy
\title{RecastVLA: From Past Interaction to Future Control with Adaptive Policy States}
\author{%
Wenbo Li$^{1}$ \quad Jun Yang$^{2}$ \quad Yiteng Chen$^{1}$ \quad Wei Zhang$^{1}$ \quad Qingyao Wu$^{1,*}$\\[0.5em]
\normalfont $^{1}$School of Software Engineering, South China University of Technology\\
\normalfont Guangzhou, China\\
\normalfont $^{2}$Yuanwu Technology, Shenzhen, China\\[0.3em]
\normalfont $^{*}$Corresponding author: \texttt{qyw@scut.edu.cn}
}
\hypersetup{
  pdftitle={RecastVLA: From Past Interaction to Future Control with Adaptive Policy States},
  pdfauthor={Wenbo Li, Jun Yang, Yiteng Chen, Wei Zhang, Qingyao Wu},
  pdfsubject={Preprint}
}

\newcommand{\method}{RecastVLA}
\newcommand{\sg}{\operatorname{sg}}
\newcommand{\LN}{\operatorname{LN}}
\newcommand{\E}{\mathbb{E}}
\newcommand{\R}{\mathbb{R}}
\definecolor{toponebg}{RGB}{169,208,142}
\definecolor{toptwobg}{RGB}{190,219,160}
\newcommand{\topone}[1]{\cellcolor{toponebg}\textbf{#1}}
\newcommand{\toptwo}[1]{\cellcolor{toptwobg}#1}
\definecolor{stateblue}{HTML}{356C9B}
\definecolor{writeorange}{HTML}{BB733C}
\definecolor{computegray}{HTML}{52616B}
\definecolor{panelgray}{HTML}{F4F6F8}
\tikzset{
  flow/.style={-{Latex[length=1.8mm]},line width=.7pt,draw=computegray},
  readflow/.style={flow,draw=stateblue},
  writeflow/.style={flow,draw=writeorange,dashed},
  box/.style={draw=computegray!65,rounded corners=2pt,fill=white,
    align=center,inner sep=4pt,minimum height=7mm},
  state/.style={box,draw=stateblue,fill=stateblue!9},
  writebox/.style={box,draw=writeorange,fill=writeorange!9},
  paneltitle/.style={font=\small\bfseries,anchor=west},
}

\begin{document}
\maketitle
% A preprint has no conference acceptance or review-status header.
\lhead{}
\renewcommand{\headrulewidth}{0pt}
\begin{abstract}
Sequential manipulation requires a robot to track what has already happened, even when the current scene no longer reveals it. Policies with explicit history representations make past interactions available as context for current decisions. We ask how action generation itself can form a persistent state for subsequent control. Building on action-side test-time training, \method{} maintains an adaptive policy state within a flow-matching vision-language-action policy. The state is represented by shared fast weights and remains fixed throughout action generation. Depth-specific interfaces read the same state, while features across depths and flow evaluations jointly define one update for the next policy call. Subsequent action losses train the initialization, interfaces, and update rule by differentiating through earlier state transitions. At deployment, updates use the policy's own action-generation features without expert action labels. Across LIBERO, RoboTwin, RoboDojo, and twelve real-robot tasks, \method{} improves mean success over a matched policy trained without test-time training, including $10.68$ percentage points on RoboTwin Clean$\to$Clean. In controlled RoboTwin comparisons, retaining state improves success, and the shared design exceeds independently trained layer-local TTT by $2.58$ points.

\end{abstract}
\section{Introduction}
\label{sec:intro}

Knowing how to press a button is different from knowing whether it has already been pressed in the current task. A momentary button can look nearly identical before and after activation, yet the next correct action depends on that earlier event~\citep{shi2025memoryvla}. Occlusion and partially completed instructions create a similar need to connect current observations with earlier events. For vision-language-action (VLA) policies, this need extends beyond learning reusable manipulation skills~\citep{kim2024openvla,black2024pi0,bjorck2025groot}. We refer to retaining the information needed to continue an ongoing task as \emph{execution memory}.

History-conditioned policies address this problem through retrieved observations, compressed context, or recurrent representations. MemoryVLA and MEM show how perceptual and semantic memories can support control that depends on earlier events~\citep{shi2025memoryvla,torne2026mem}. These designs determine what information from past interaction is available for control. We focus on a complementary question: how should action generation transform that experience into a persistent policy state that guides subsequent decisions?

\begin{figure}[t]
  \centering
  \includegraphics[width=\linewidth]{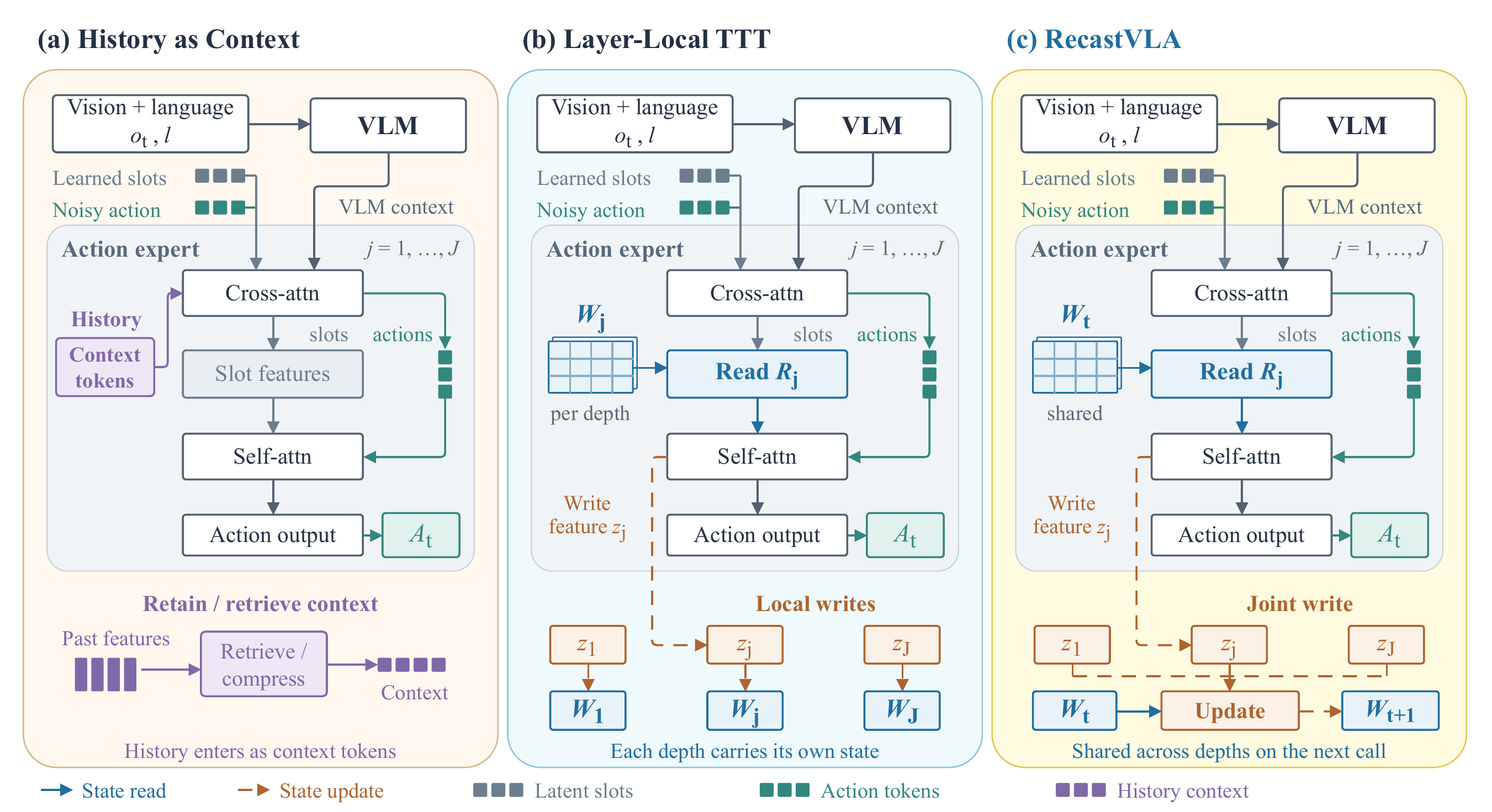}
  \caption{\textbf{Three routes from history to action generation.} (a) Retrieved or compressed history provides context tokens. (b) Layer-local TTT maintains independent fast weights at each depth. (c) \method{} reads one shared state through depth-specific interfaces and combines their features into a single update per policy call.}
  \label{fig:teaser}
\end{figure}

Action generation is where a policy turns perception and language into a control decision. A generative action expert jointly processes visual-language features and candidate actions, producing information tied to the decisions being formed. Test-time training (TTT) represents history in the weights of a small model updated as a sequence arrives~\citep{sun2024ttt}. RoboTTT applies this principle inside an action expert and learns fast-weight updates through trajectory supervision~\citep{jiang2026robottt}. In layer-local recurrence, each depth directly reads and writes its own fast weights. We study whether these specialized computations benefit from jointly forming and accessing a common state (Figure~\ref{fig:teaser}).

\method{} maintains an \emph{adaptive policy state} in the weights of a small model shared across depths. Layer-specific interfaces query the model in their own feature coordinates. All action computations within a call read a fixed incoming state; their features jointly define one update for the next call. Information written at one depth can therefore influence reads at other depths on subsequent calls. Differentiating later action losses through these state transitions trains the initialization, interfaces, and update rule. At deployment, only the state changes, using the policy's own action-generation features without expert action labels.

\method{} extends action-side TTT with one state that action-expert depths jointly read and update. We evaluate the complete policy on LIBERO, RoboTwin, and RoboDojo, then use a fixed checkpoint to examine the effects of state continuity and imported history. An independently trained layer-local model tests the shared organization with the same per-state architecture and update timing. Twelve real-robot tasks further evaluate progress tracking and ordering constraints, while revealing remaining geometric failures.

\section{Related Work}
\label{sec:related}

\paragraph{Long-Context Robot Policies.}
Long-context policies retain earlier experience through retrieval and compressed representations. MemoryVLA and Zeva retrieve memory tokens and interaction context, respectively~\citep{shi2025memoryvla,chen2026zeva}, while MEM combines compressed recent video with longer-term language memory~\citep{torne2026mem}. Temporal prediction provides a complementary training signal: Past-Token Prediction regularizes diffusion policies to model dependencies between earlier and later actions~\citep{torne2025ptp}. \method{} focuses on how action-generation features update the policy's state for subsequent control.

\paragraph{Test-Time Training.}
Early TTT adapts predictors through self-supervision on unlabeled test inputs under distribution shift~\citep{sun2020ttt}. TTT sequence models incorporate this adaptation into recurrent computation, with a learned model serving as the hidden state~\citep{sun2024ttt}. In robotics, RoboTTT places fast-weight layers after attention in a diffusion action expert and learns their update dynamics through sequential flow-matching losses~\citep{jiang2026robottt}. It is the closest precedent for our action-side recurrence and supervision. WAM-TTT adapts video-side memory to steer a world-action model~\citep{feng2026wamttt}, while VITA targets value estimation~\citep{ziakas2026vita}. \method{} combines depth-specific access with one shared state, keeping it fixed throughout action generation and updating it once for the next policy call.

\paragraph{Shared Recurrent State Across Network Depth.}
Feedback Transformer makes earlier representations available across network depths~\citep{fan2020feedback}. Shared Global Workspace studies communication among specialized modules through a common bottleneck~\citep{goyal2022workspace}. Universal Test-Time Training (uTTT) describes globally shared fast-weight modules with updates aggregated across access sites~\citep{cai2026uttt}. These works provide precedents for cross-depth sharing and aggregate writes. Together with action-side TTT, they motivate our study of shared execution memory inside iterative robot action generation. In \method{}, subsequent action losses train a joint state update from multiple depths and flow evaluations.

\section{Learning Execution Memory through Adaptive Policy States}
\label{sec:method}

\begin{figure}[!htbp]
  \centering
  \includegraphics[width=\linewidth]{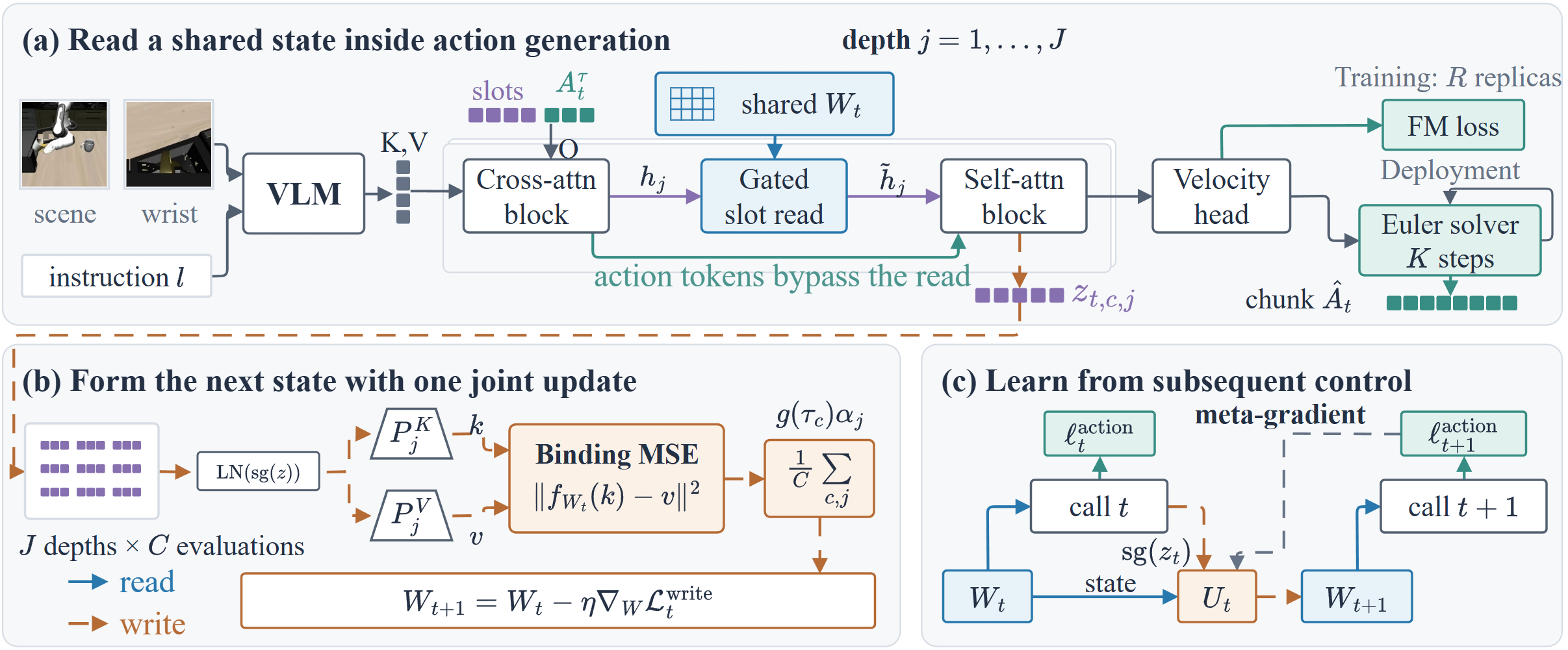}
  \caption{\textbf{\method{} overview.} (a) Depth-specific slot interfaces read a shared state held fixed throughout action generation. (b) Features across depths and flow evaluations define one joint update for the next call. (c) Subsequent action losses train the initialization, interfaces, and update rule through these state transitions.}
  \label{fig:method}
\end{figure}

Execution memory denotes the function of retaining information needed for continued control. \method{} implements it as an \emph{adaptive policy state}, represented by fast weights shared across the depths of a generative action expert~\citep{chi2023diffusion,black2024pi0}. Each call reads its incoming state, generates an action chunk, and uses internal features to update the state for the next call (Figure~\ref{fig:method}).

\subsection{Adaptive Policy State for Sequential Control}
\label{sec:formulation}

Let $o_t$ denote the current visual observation, $l$ the language instruction, and $A_t\in\R^{H\times d_a}$ an action chunk. The policy has slow parameters $\theta$ for the visual-language backbone and action expert, and $\psi$ for the memory interfaces and update rule. Its persistent state $W_t$ comprises the weights and biases of a two-layer MLP $f_{W_t}$ with GELU activation. Each policy call reads this state to generate an action chunk and collect internal features $\mathcal Z_t$:
\begin{equation}
  (\hat A_t,\mathcal Z_t)=\pi_{\theta,\psi}(o_t,l;W_t),
  \qquad W_{t+1}=U_\psi(W_t,\mathcal Z_t)
  \label{eq:recurrence}
\end{equation}
All action computations within a call use $W_t$; the update affects subsequent calls. A new episode starts from the learned initialization $W_0$. The index $t$ orders policy calls, while $\tau\in[0,1]$ denotes flow time within action generation.

At deployment, only $W_t$ changes; $\theta$ and $\psi$ remain fixed. Training forms write features from noisy expert action chunks, whereas deployment uses features produced during iterative action generation. The update precedes physical execution of the generated chunk and does not require feedback on its outcome.

\subsection{Specialized Access to Shared Execution Memory}
\label{sec:access}

The action expert alternates cross-attention to visual-language features with self-attention over learned latent slots and action tokens. The slots provide a compact interface to the state. Let~$h_{t,c,j}\in\R^{N\times D}$ denote the features of $N$ slots of width $D$ after cross-attention at access depth $j$ and flow evaluation $c$, with flow time $\tau_{t,c}$. The state read adds a gated residual to the slot features:
\begin{equation}
  \tilde h_{t,c,j}
  =h_{t,c,j}+\tanh(\gamma_j)\,P^O_j
    f_{W_t}\!\left(P^Q_j\LN(h_{t,c,j})\right)
  \label{eq:read}
\end{equation}
The linear maps act token-wise: $P^Q_j$ projects into the memory space and $P^O_j$ maps back, while the learned scalar $\gamma_j$ gates the residual. We use $N=32$ slots of width $D=768$ at $J=8$ access sites, with a $256\!\to\!512\!\to\!256$ fast MLP. Action tokens bypass the state read and interact with the adapted slots in the next self-attention block. Let $z_{t,c,j}$ denote the slot features after that block.

The interfaces let different depths query the same history-dependent function in their own feature coordinates. Information written at depth $j$ can thus affect reads at depth $j'$ on a later call. Section~\ref{sec:ablation} compares this shared organization with independent layer-local states.

\subsection{Forming Memory from Action-Generation Features}
\label{sec:write}

To write the current call's features, each site constructs learned keys and values from its post-self-attention slots:
\begin{equation}
  k_{t,c,j}=P^K_j\LN(\sg(z_{t,c,j})),\qquad
  v_{t,c,j}=P^V_j\LN(\sg(z_{t,c,j}))
  \label{eq:kv}
\end{equation}
The stop-gradient operator $\sg$ treats source features as data for adaptation. The inner objective fits the fast model to these learned key--value associations. For $C$ flow evaluations, $J$ access depths, and memory dimension $M$,
\begin{equation}
  \mathcal L^{\mathrm{write}}_t(W)=
  \frac{1}{C}\sum_{c=1}^{C}g_\psi(\tau_{t,c})
    \sum_{j=1}^{J}\alpha_j
    \frac{\|f_W(k_{t,c,j})-v_{t,c,j}\|_F^2}{NM},
  \qquad \alpha_j=\frac{e^{\beta_j}}{\sum_{j'}e^{\beta_{j'}}}
  \label{eq:write_loss}
\end{equation}
The learned function $g_\psi(\tau)\in(0,1)$ is a two-layer MLP with hidden width $16$, SiLU activation, and a sigmoid output; $\alpha_j$ weights access sites. Flow-time weights are not normalized across evaluations, so they control both relative contributions and overall update magnitude.

We evaluate all terms at the incoming state and take a single gradient step,
\begin{equation}
  W_{t+1}=W_t-\eta\left.
      \nabla_W\mathcal L^{\mathrm{write}}_t(W)\right|_{W=W_t},
  \qquad \eta=\eta_0\exp(\rho)
  \label{eq:update}
\end{equation}
The learned scalar $\rho$ controls a positive step size, with $\eta_0=0.1$. The transition lies at the policy-call boundary because depths and flow evaluations are internal computations of the same action-generation process. They read a common incoming state and jointly define one update after action generation. This separates within-call computation from across-call adaptation: $W_t$ conditions the complete current decision, while features from that computation determine $W_{t+1}$ for the next decision. Subsequent action losses train the write projections and update rule through the resulting state's effect on later predictions.

\subsection{Learning from Subsequent Control}
\label{sec:learning}

\paragraph{Sequential action supervision.}
Training samples an ordered window of $T=4$ observations from one demonstration episode. Each window begins at $W_0$. At frame $t$, we draw $R$ independent noise/time samples and form flow-matching inputs and velocity targets,
\begin{equation}
  A^{\tau}_{t,r}=(1-\tau_{t,r})\epsilon_{t,r}+\tau_{t,r}A_t^*,
  \qquad u^*_{t,r}=A_t^*-\epsilon_{t,r},
  \quad\epsilon_{t,r}\sim\mathcal N(0,I)
  \label{eq:fm_input}
\end{equation}
We sample $b\sim\mathrm{Beta}(1.5,1)$ and set $\tau=(s-\min(b,s))/s$ with $s=0.999$. The velocity predictor $v_{\theta,\psi}$ reads the same incoming $W_t$ for all $R$ replicas. Their features jointly define one update with $C=R$. We omit the final state update because no later loss in the training window depends on the resulting state. All frames contribute to the outer objective for $\Theta=(\theta,\psi,W_0)$:
\begin{equation}
  \begin{aligned}
  \mathcal L^{\mathrm{action}}(\Theta)=
  &\E_{\mathcal D,\tau,\epsilon}\!\left[
  \frac{1}{TR}\sum_{t=0}^{T-1}\sum_{r=1}^{R}
    \ell_{t,r}(W_t)\right],\\
  \ell_{t,r}(W_t)=
  &\frac{\|v_{\theta,\psi}(o_t,l,A^{\tau}_{t,r},\tau_{t,r};W_t)
               -u^*_{t,r}\|_F^2}{Hd_a}
  \end{aligned}
  \label{eq:outer}
\end{equation}
An action loss at frame $t+1$ backpropagates through Eq.~\eqref{eq:update} to the write projections and update parameters used at frame $t$. Gradients through the unrolled states also train $W_0$ and earlier updates within the window. The stop-gradient in Eq.~\eqref{eq:kv} preserves gradients to the key/value projections and fast-weight update while blocking gradients into the source features through the write branch. The visual-language backbone and action expert remain trained by the action objective. Algorithm~\ref{alg:train} summarizes this procedure.

\paragraph{Deployment.}
The policy starts each call from Gaussian action noise and integrates the learned velocity field with $K$ Euler evaluations, all conditioned on $W_t$. Features from these evaluations jointly define one update with $C=K$ after action generation. The next call reads $W_{t+1}$. Deployment requires only gradients with respect to the fast weights; no computational graph is retained across calls.

Training uses adjacent recorded frames, whereas deployment advances by executed action chunks. The LIBERO recipe uses $H=8$ actions and $K=4$ Euler evaluations, requesting a new observation after all eight actions execute.

\begin{algorithm}[!htb]
\caption{Training \method{} through subsequent control}
\label{alg:train}
\begin{algorithmic}[1]
\Require Ordered episode window $\{(o_t,l,A_t^*)\}_{t=0}^{T-1}$; slow parameters $\Theta$
\State $W\gets W_0$; $\mathcal L\gets0$ \Comment{One independent state per window}
\For{$t=0,\ldots,T-1$}
  \State Sample $R$ independent $(\epsilon_{t,r},\tau_{t,r})$; construct $A^\tau_{t,r},u^*_{t,r}$
  \State Predict velocities and collect $z_{t,r,j}$, reading fixed $W$ at every site
  \State $\mathcal L\gets\mathcal L+\frac{1}{TR}\sum_{r=1}^{R}\ell_{t,r}(W)$
  \If{$t<T-1$}
    \State Form $k,v$ from stopped source features using Eq.~\eqref{eq:kv}
    \State $W\gets W-\eta\nabla_W\mathcal L^{\mathrm{write}}_t(W)$ \Comment{Retain meta-gradients}
  \EndIf
\EndFor
\State Update $\Theta$ using $\nabla_\Theta\mathcal L$
\end{algorithmic}
\end{algorithm}

\section{Experiments}
\label{sec:experiments}

We evaluate task success across three simulation benchmarks, then use controlled comparisons on RoboTwin to examine state continuity, imported history, and cross-depth sharing. Twelve real-robot tasks assess performance and failure modes in physical execution.

\paragraph{Model and matched baseline.}
Our implementation builds on the StarVLA/VLAct codebase~\citep{community2026starvla,yang2026vlact}. We initialize the visual-language backbone from Qwen3-VL-4B-Instruct and train it jointly with a newly initialized GR00T-style action expert and TTT modules. Matched Base is trained independently with TTT disabled, retaining the four-frame action objective. Appendix~\ref{app:evaluation} reports the training data, update budgets, and evaluation counts.

\subsection{Main Results}
\label{sec:main_results}

Table~\ref{tab:main} reports task success on three simulation benchmarks. LIBERO averages its Spatial, Object, Goal, and Long suites with equal weight and reports Long separately~\citep{liu2023libero}. RoboTwin evaluates policies trained on clean demonstrations under Clean$\to$Clean and Clean$\to$Random conditions~\citep{chen2025robotwin2}. RoboDojo reports full-task success separately for Long-Horizon and Memory tasks~\citep{chen2026robodojo}. Together, these settings cover broad manipulation performance, distribution shift, and tasks that require information from earlier interactions.

We compare with $\pi_{0.5}$, X-VLA, and Spatial Forcing~\citep{pi2025pi05,zheng2025xvla,li2025spatialforcing}, as well as ABot-M0, Fast-WAM, and OpenWAM-$\alpha$~\citep{yang2026abotm0,yuan2026fastwam,wang2026openwam}. Table~\ref{tab:main} distinguishes these published results from our matched comparison in the lower block.

% Benchmark scores supplied by the author.
% Preserve each row's supplied decimal precision. Rank distinct numeric
% scores across all policies within each column.
\begin{table}[!htb]
\centering
\caption{\textbf{Task success rates (\%; higher is better).} LIBERO Avg weights four suites equally. RoboTwin C$\to$C and C$\to$R denote Clean$\to$Clean and Clean$\to$Random. The lower block compares Matched Base with \method{}. Darker and lighter green mark the best and second-best distinct scores per column.}
\label{tab:main}
\small
\setlength{\tabcolsep}{3.3pt}
\renewcommand{\arraystretch}{1.15}
\begin{tabular*}{\linewidth}{@{\extracolsep{\fill}}lrrrrrr@{}}
\toprule
 & \multicolumn{2}{c}{LIBERO} & \multicolumn{2}{c}{RoboTwin} & \multicolumn{2}{c}{RoboDojo}\\
\cmidrule(lr){2-3}\cmidrule(lr){4-5}\cmidrule(lr){6-7}
Policy & Avg & Long & C$\to$C & C$\to$R & Long-Horizon & Memory\\
\midrule
$\pi_{0.5}$ & 96.9 & 92.4 & 70.7 & \toptwo{46.0} & 14.67 & 4.56\\
X-VLA & 98.1 & \toptwo{97.6} & 68.0 & 20.9 & 9.75 & 3.56\\
Spatial Forcing & 98.5 & 96.0 & 77.2 & 26.7 & 14.58 & 4.11\\
ABot-M0 & 98.6 & 96.6 & 70.7 & 36.0 & 0.50 & 2.22\\
Fast-WAM & 97.6 & 95.2 & \toptwo{77.8} & 1.9 & 5.17 & 3.44\\
OpenWAM-$\alpha$ & \topone{99.3} & \topone{98.2} & \topone{89.4} & \topone{48.7} & \topone{25.33} & \topone{9.11}\\
\midrule
Matched Base & 96.5 & 92.0 & 61.7 & 10.5 & 6.50 & 3.33\\
\textbf{\method{}} & \toptwo{98.80} & 96.40 & 72.38 & 19.24 & \toptwo{16.75} & \toptwo{8.67}\\
\bottomrule
\end{tabular*}
\end{table}

\begin{samepage}
\method{} improves success over Matched Base in all six reported settings, including $10.68$ percentage points on RoboTwin Clean$\to$Clean (\mbox{Table~\ref{tab:main}}). On LIBERO, the gain is $2.30$ points overall and $4.40$ points on Long. Improvements extend to RoboDojo's Long-Horizon and Memory categories, with approximately $5.3$ points gained on Memory. On RoboTwin Clean$\to$Random, success improves over Matched Base but remains substantially below Clean$\to$Clean. These results assess the complete architecture; the following controls examine state continuity, imported history, and cross-depth organization under Clean conditions.
\par
\end{samepage}

\subsection{Does Accumulated History Improve Control?}
\label{sec:memory_analysis}

Table~\ref{tab:intervention} compares four state conditions on all 50 RoboTwin Clean$\to$Clean tasks using the same checkpoint. Full versus Reset tests within-episode continuity; Similar and Other additionally probe sensitivity to adaptation from another task. All conditions share initial scenes, environment seeds, sampling randomness, and episode budgets. We average metrics equally over target tasks.

\paragraph{State continuity.}
\emph{Full} follows the main evaluation: each episode starts at $W_0$, and every call reads the current state, generates an action chunk, and updates the state for the next call. \emph{Reset} restores the learned $W_0$ before every call, preventing history from accumulating. Reset retains the trained state-read interfaces of the same checkpoint; it is a deployment intervention, whereas Matched Base is trained independently without TTT.

\paragraph{Experience at initialization.}
\emph{Similar} selects another task in the target's manipulation group; \emph{Other} selects a task outside that group. Appendix~\ref{app:taskgroups} lists the ten groups. Both initialize the target with the final state accumulated from $W_0$ over a complete source rollout. This import occurs once, before the target's first action chunk; normal state updates then continue throughout the target episode.

\begin{figure}[t]
  \centering
  \includegraphics[width=\linewidth]{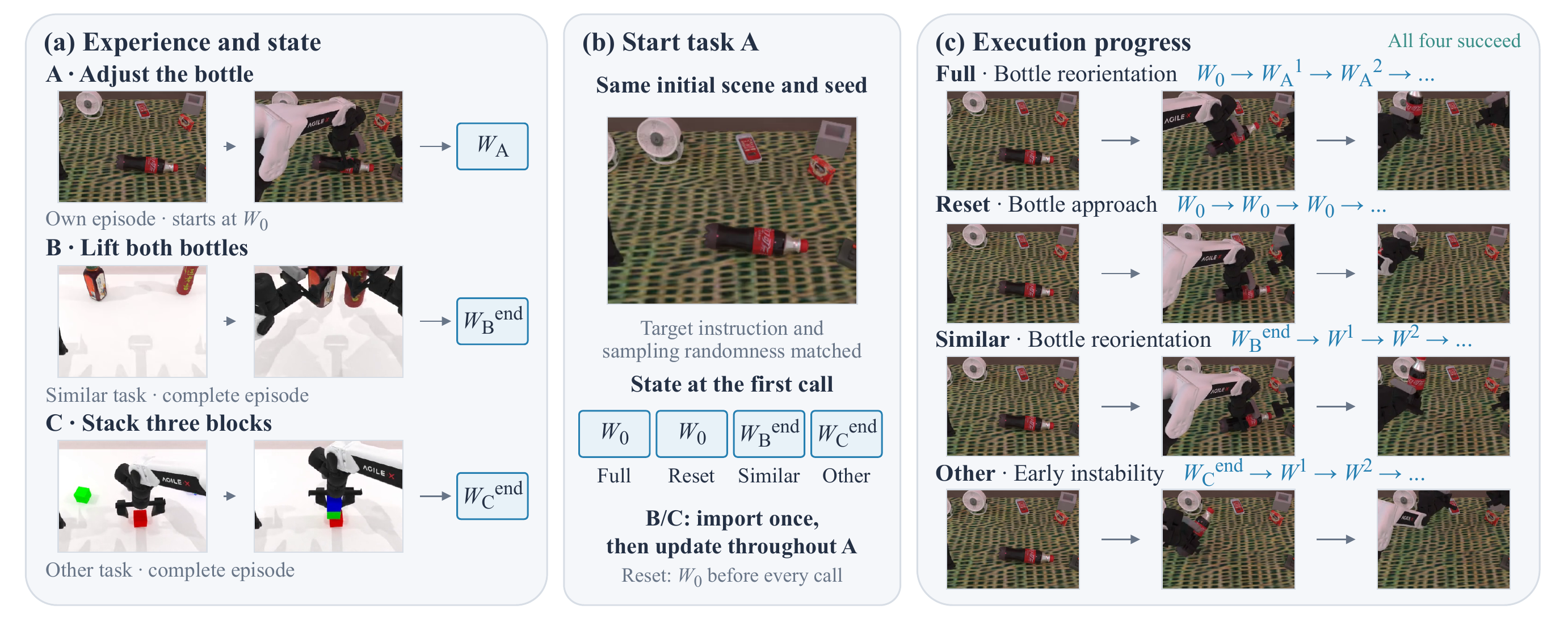}
  \caption{\textbf{State interventions in a matched scene.} (a) Task histories and accumulated states. (b) Target-scene initialization. (c) All four conditions eventually succeed despite differences in intermediate behavior. Keyframes run left to right.}
  \label{fig:state_intervention}
\end{figure}

\paragraph{Closed-loop performance.}
Success rate (SR) and normalized completion steps (NCS; failures score one) measure performance over complete target episodes; Appendix~\ref{app:state_metrics} defines both metrics. Figure~\ref{fig:state_intervention} compares bottle-adjustment executions, using bottle lifting and block stacking as similar- and other-task histories. The selected rollouts all succeed but exhibit different intermediate behaviors; the Other rollout shows early instability followed by recovery.

Full exceeds Reset by $4.68$ percentage points in SR (Table~\ref{tab:intervention}). Successful trials in both conditions use nearly the same mean fraction of the episode budget, about $30\%$; the lower NCS for Full therefore primarily reflects fewer failures. These fractions are computed over each condition's successful trials, which need not comprise the same episodes.

\begin{samepage}
Similar raises SR over Full by $2.22$ percentage points, whereas Other lowers it by $11.98$ points (Table~\ref{tab:intervention}). NCS is lower for Similar and higher for Other; successful Other trials use about $48\%$ of the budget. Because state updates continue throughout the target episode, these comparisons measure the effect of initialization together with continued adaptation to the target.
\par
\end{samepage}

Other also underperforms Reset by $7.3$ percentage points ($60.4\%$ versus $67.7\%$). Reset prevents accumulated history from carrying across calls, whereas Other begins from a state formed by a task in another manipulation group and then resumes ordinary target-task updates. Thus, carrying an adapted state is not beneficial by itself: its source matters. In this evaluation, subsequent updates do not fully offset the adverse initialization in overall rollout success. Together with Full versus Reset, this comparison shows why state continuity and the suitability of prior experience both matter for control.

% Compact, independently numbered tables; keep text at its native font size.
\begin{table}[!htbp]
  \centering
  \begin{minipage}[t]{0.53\linewidth}
    \vspace{0pt}
    % Author-reported state metrics; Full SR is reused from the main table.
\caption{\textbf{State controls on RoboTwin C$\to$C.} SR (\%) and NCS averaged over all 50 target tasks. Full SR is reused from Table~\ref{tab:main}.}
\label{tab:intervention}
\centering
\small
\setlength{\tabcolsep}{3pt}
\renewcommand{\arraystretch}{1.17}
\begin{tabular}{@{}lcc@{}}
\toprule
State rule & SR $\uparrow$ & NCS $\downarrow$\\
\midrule
Full & 72.38 & 0.4945\\
Reset each call & 67.7 & 0.5273\\
Similar-task init. & 74.6 & 0.4719\\
Other-task init. & 60.4 & 0.6831\\
\bottomrule
\end{tabular}

  \end{minipage}\hfill
  \begin{minipage}[t]{0.44\linewidth}
    \vspace{0pt}
    % Included in the right minipage of the paired experiment-table float.
\caption{\textbf{State organization on RoboTwin.} C$\to$C success (\%) at a fixed per-state MLP design.}
\label{tab:ablation}
\centering
\small
\setlength{\tabcolsep}{3pt}
\renewcommand{\arraystretch}{1.17}
\begin{tabular}{@{}llr@{}}
\toprule
Policy & State & SR $\uparrow$\\
\midrule
Matched Base & None & 61.7\\
Layer-local TTT & Per depth & 69.8\\
\textbf{\method{}} & Shared & 72.38\\
\bottomrule
\end{tabular}

  \end{minipage}
\end{table}

\begin{samepage}
\subsection{How Does State Sharing Affect Control?}
\label{sec:ablation}

Table~\ref{tab:ablation} compares independently trained Matched Base, layer-local TTT, and \method{} on RoboTwin Clean$\to$Clean. The layer-local control assigns each access site its own state, learned initialization, step size, and flow-time gate. Each state updates only from features at its corresponding depth. The control retains our feature sources, access interfaces, sequential action objective, and call-boundary timing; Appendix~\ref{app:local_control} gives its update.
\par
\end{samepage}

\begin{samepage}
Each fast model has the same architecture, giving the layer-local control eight times the persistent-state capacity. Under this per-state design, \method{} improves success over layer-local TTT by $2.58$ percentage points ($72.38\%$ versus $69.8\%$). The observed advantage therefore cannot be attributed to giving \method{} more persistent-state capacity; it achieves higher success with fewer fast-state values in this comparison. The $10.68$-point gain over Matched Base measures the benefit of the complete TTT architecture. Matched Base and \method{} scores are reused from Table~\ref{tab:main}.
\par
\end{samepage}

\subsection{Real-Robot Evaluation}
\label{sec:real_robot}

\begin{figure}[!htbp]
  \centering
  \includegraphics[width=\linewidth]{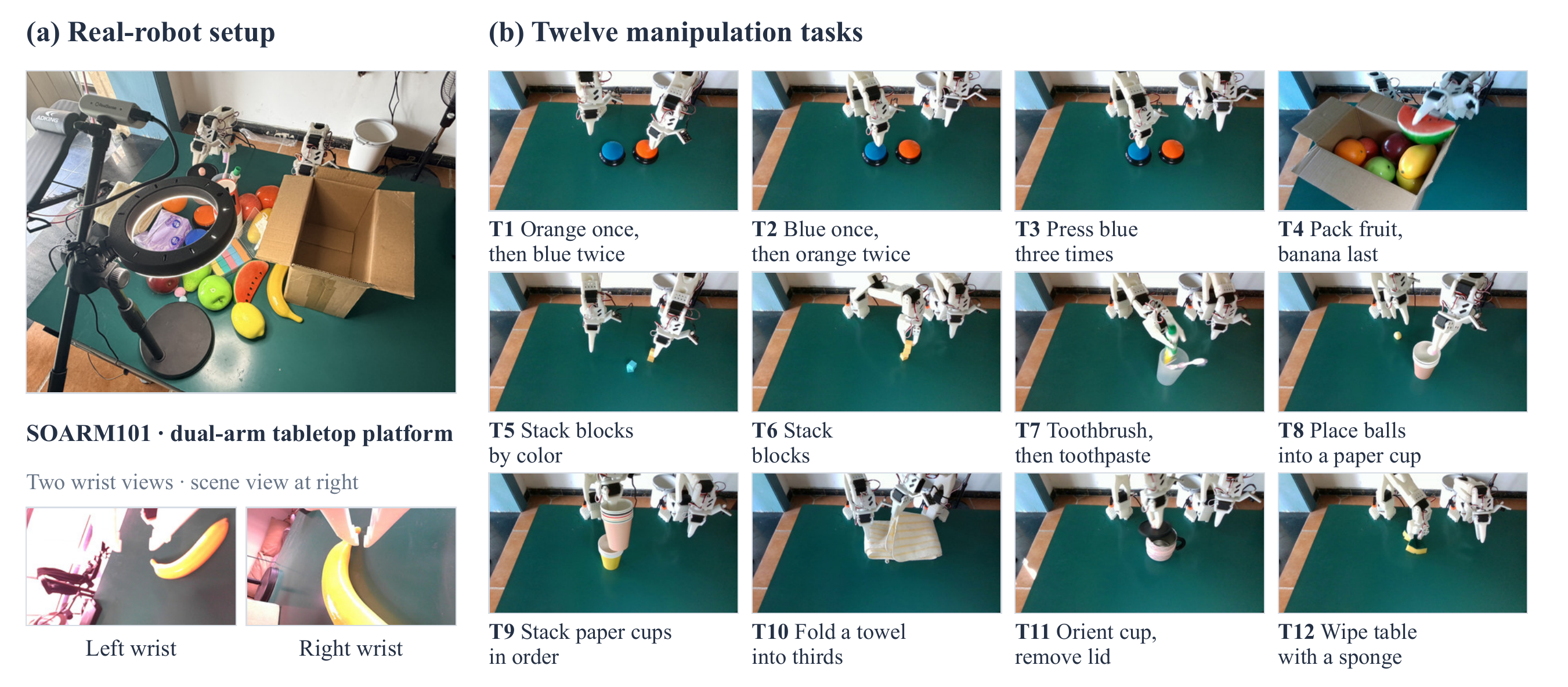}
  \caption{\textbf{Real-robot setup and tasks.} (a) Dual-arm platform with LRCP G720P wrist cameras. (b) The twelve evaluation tasks.}
  \label{fig:real_overview}
\end{figure}

We compare $\pi_{0.5}$, Matched Base, and \method{} on twelve tasks using a dual-arm SOARM101 platform (Figure~\ref{fig:real_overview}). All three policies are fine-tuned on the same $55$ training tasks with $10$ demonstrations per task ($550$ in total), spanning nine task families, under the same protocol. The twelve evaluation tasks are held-out variants involving new operation sequences or combinations and new scene or object configurations relative to the fine-tuning data. Table~\ref{tab:real} shows six representative tasks and the average over all twelve. Appendix~\ref{app:real} gives the training composition, success criteria, and complete counts. Across $240$ trials per policy, \method{} succeeds in $131$ ($54.58\%$), compared with $102$ ($42.50\%$) for $\pi_{0.5}$ and $74$ ($30.83\%$) for Matched Base.

\begin{table}[!htbp]
\centering
\caption{\textbf{Real-robot task success (\%; higher is better).} Each policy is evaluated on 20 trials per task. Six representative tasks are shown; Avg gives equal weight to all twelve tasks in Figure~\ref{fig:real_overview}.}
\label{tab:real}
\small
\setlength{\tabcolsep}{3pt}
\renewcommand{\arraystretch}{1.15}
\begin{tabular*}{\linewidth}{@{\extracolsep{\fill}}lccccccc@{}}
\toprule
Policy & \shortstack{Button\\sequence\\(T1)} & \shortstack{Repeated\\pressing\\(T3)} & \shortstack{Fruit\\packing\\(T4)} & \shortstack{Color\\stacking\\(T5)} & \shortstack{Brush \&\\paste\\(T7)} & \shortstack{Towel\\folding\\(T10)} & \shortstack{Avg\\(12 tasks)\\$\uparrow$}\\
\midrule
$\pi_{0.5}$ & 30 & 15 & 45 & 50 & 45 & 30 & 42.50\\
Matched Base & 20 & 5 & 35 & 45 & 20 & 20 & 30.83\\
\textbf{\method{}} & 75 & 60 & 60 & 60 & 35 & 40 & 54.58\\
\bottomrule
\end{tabular*}
\end{table}

\begin{samepage}
\paragraph{Task-level gains.}
Relative to $\pi_{0.5}$, the advantage is concentrated in button sequencing and repeated pressing (T1--T3), with $66.67\%$ success versus $23.33\%$. These three tasks account for $26$ of the $29$ net additional successes across the full evaluation. The remaining nine tasks average $50.56\%$ versus $48.89\%$; toothbrush-and-toothpaste placement, cup orientation and lid removal, and wiping still favor $\pi_{0.5}$. Compared with Matched Base, success increases by $50$ percentage points on T1--T3 and by $15$ points on T4--T12 (Table~\ref{tab:real_full}).
\par
\end{samepage}

The button tasks are distinguished by weakly observable progress. The same buttons remain available throughout a trial, but the next required press depends on the order and number of earlier presses. A successful policy must continue from its current task progress even when the visible objects change little. The concentration of gains on these tasks is consistent with the role of an adaptive policy state in carrying information between successive control decisions.

\begin{samepage}
\paragraph{Execution behavior.}
In supplementary executions, the robot preserves the banana-last constraint after fruit and box movements and resumes placement after a stack collapses. These executions, including the T4 perturbations, are separate from the scored trials. Observed failures include extra button presses, mixed-color stacking, cup-rim misalignment, and unstable stacks (Appendix~\ref{app:real}).
\par
\end{samepage}

These rollouts distinguish retaining task order from achieving accurate placement: in T7, the robot handles the objects in the required order but leaves the toothbrush across the cup rim.

\begin{samepage}
\section{Conclusion}
\label{sec:conclusion}

\method{} forms execution memory by updating shared fast weights from action-generation features. Subsequent action losses train the initialization, interfaces, and update rule. The policy improves mean success over Matched Base across three simulation benchmarks and twelve real-robot tasks. Retaining state improves within-episode control on RoboTwin, while imported-state probes show that the effect of prior adaptation depends on its source. The shared design also achieves higher success than its layer-local counterpart under the tested per-state architecture.
\par
\end{samepage}

\paragraph{Limitations.}
Training uses four adjacent demonstration frames, whereas deployment advances through successive action chunks and forms write features from the policy's own action generation. A shared state can also permit interference between feature sources, and stopping gradients at the write features restricts temporal credit assignment. Progress-tracking and geometric errors persist in real-robot execution, including extra button presses and cup-rim misalignment.

\paragraph{Future work.}
An open question is how shared adaptive states should be shaped when training spans longer portions of an interaction. Training with deployment-like temporal spacing and policy-generated write features could reveal how later action losses influence earlier state updates over longer horizons. A complementary question is which information should continue to affect control within a fixed shared state, for how long, and how that influence should change as a task progresses.

\section*{AI Disclosure}

Generative AI tools were used for language polishing of selected sentences. All scientific content, analyses, and reported results were reviewed and verified by the authors.

% Keep the reference list on a separate page after the main text.
\clearpage
\bibliography{refs/references}
\bibliographystyle{iclr2027_conference}
\clearpage
\appendix
% Appendix-only spacing: keep full-size sequence images close to their captions.
\setlength{\intextsep}{6pt plus 2pt minus 2pt}
\setlength{\abovecaptionskip}{6pt}
\setlength{\belowcaptionskip}{0pt}
% Keep appendix float pages top-aligned instead of vertically centered.
\makeatletter
\setlength{\@fptop}{0pt}
\makeatother
\section{Implementation and Evaluation Details}
\label{app:memory}
\label{app:experiments}

\subsection{Training Data, Configuration, and Evaluation Scope}
\label{app:evaluation}

Table~\ref{tab:experiment_setup} summarizes the training data and evaluation protocol. LIBERO training combines Spatial, Object, Goal, and Long with $432$, $454$, $428$, and $379$ demonstrations, respectively. RoboTwin uses $50$ clean demonstrations per task; randomized scenes are used only for evaluation. RoboDojo uses all $3{,}500$ demonstrations from the $35$ tasks in the training release.

\begin{table}[!htbp]
\centering
\caption{\textbf{Simulation training data and evaluation scope.} Demonstrations denote recorded training episodes, not sampled windows. Evaluations use one run per setting. LIBERO Long is included in Avg; RoboTwin uses the same clean-trained policy for both evaluation settings. $R$ is the number of flow-matching replicas per training observation.}
\label{tab:experiment_setup}
\small
\setlength{\tabcolsep}{5pt}
\renewcommand{\arraystretch}{1.2}
\begin{tabularx}{\linewidth}{@{}lrrXr@{}}
\toprule
Benchmark & Train tasks & Demonstrations & Evaluation tasks $\times$ trials & $R$\\
\midrule
LIBERO & 40 & 1,693 & Avg: $40\times50$; Long: $10\times50$ & 4\\
RoboTwin & 50 & 2,500 & C$\to$C: $50\times100$; C$\to$R: $50\times100$ & 4\\
RoboDojo & 35 & 3,500 & Long-Horizon: $8\times50$; Memory: $6\times50$ & 8\\
\bottomrule
\end{tabularx}
\end{table}

\paragraph{Initialization and optimization.}
The visual-language backbone starts from Qwen3-VL-4B-Instruct, with a newly initialized action expert and TTT modules. Training updates the backbone, action expert, and TTT slow parameters jointly. Matched Base disables and freezes the TTT modules while optimizing the same four-frame action objective. Across benchmarks, the matched policies use $100{,}000$ optimizer updates with four windows per update, window length $T=4$, and frame stride one; the RoboTwin layer-local control uses the same budget. This gives $1.6$ million observation presentations per policy, including repeated samples. Flow-matching replicas reuse each observation and do not increase this count.

\paragraph{Access sites and action generation.}
The DiT-B expert has $16$ alternating cross- and self-attention blocks of width $768$. Its eight access sites read after each cross-attention block and collect write features after the following self-attention block. Each site uses all $32$ latent slots. Action chunks contain $8$, $32$, and $50$ actions for LIBERO, RoboTwin, and RoboDojo, respectively; deployment uses $K=4$ Euler evaluations. Training uses the benchmark-specific replica counts in Table~\ref{tab:experiment_setup}.

\paragraph{Evaluation and aggregation.}
Matched Base and \method{} use the same evaluation protocol, with one run per setting. LIBERO Avg weights its four suites equally and includes all $2{,}000$ trials; Long reports its $500$-trial subset. RoboTwin evaluates $5{,}000$ trials in each setting. RoboDojo reports $400$ Long-Horizon trials and $300$ Memory trials. We average task success rates within each evaluation category. With equal trial counts per task, these averages equal the pooled episode success rates.

\subsection{Layer-Local TTT Control}
\label{app:local_control}

The control follows a RoboTTT-style layer-local organization~\citep{jiang2026robottt}, adapted to our interfaces and call-boundary update schedule. Each of the $J=8$ access depths maintains an independent fast-weight module with its own learned initialization $W_0^{(j)}$, step size $\eta_j$, and flow-time gate $g_{\psi_j}$. It uses the same $256\!\to\!512\!\to\!256$ GELU MLP, including biases, slot sources, and read/write projection architecture as the shared-state design. Each read in Eq.~\eqref{eq:read} uses $W_t^{(j)}$ at its corresponding depth.

Keys and values follow Eq.~\eqref{eq:kv}. Since writes are not aggregated across depths, the shared-state weights $\alpha_j$ are omitted. Each module has the local objective
\begin{equation}
  \mathcal L_{t,j}^{\mathrm{local}}(W^{(j)})
  =\frac{1}{C}\sum_{c=1}^{C}g_{\psi_j}(\tau_{t,c})
    \frac{\|f_{W^{(j)}}(k_{t,c,j})-v_{t,c,j}\|_F^2}{NM},
  \label{eq:local_write_loss}
\end{equation}
and updates independently as
\begin{equation}
  W_{t+1}^{(j)}=W_t^{(j)}
  -\eta_j\left.\nabla_{W^{(j)}}\mathcal L_{t,j}^{\mathrm{local}}(W^{(j)})
       \right|_{W^{(j)}=W_t^{(j)}},
  \qquad \eta_j=\eta_0\exp(\rho_j)
  \label{eq:local_update}
\end{equation}
The gate architecture and initial step-size scale $\eta_0=0.1$ match the shared design. All modules read their incoming states throughout one policy call and apply their updates only after action generation. Training retains the four-frame objective and differentiates through the local state transitions; deployment fixes the learned initializations, gates, and step sizes while updating only the fast states.

\begin{figure}[!htbp]
  \centering
  \includegraphics[width=\linewidth]{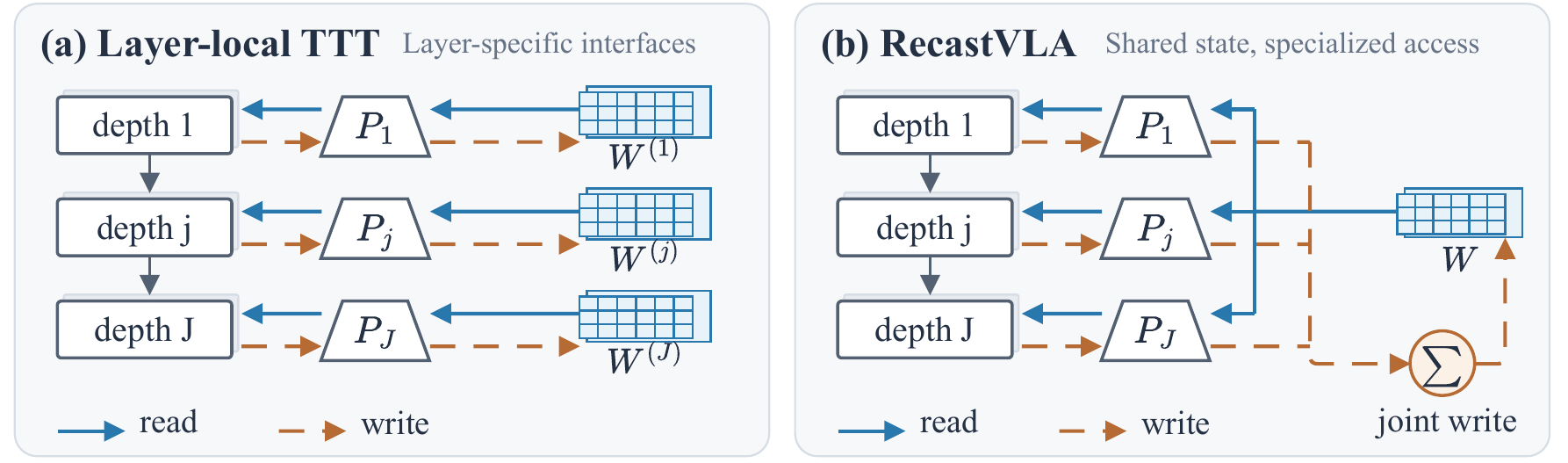}
  \caption{\textbf{Layer-local and shared adaptive states.} (a) Independent fast weights $W^{(j)}$ at each depth. (b) Depth-specific interfaces $P_j$ access a shared state $W$ with one joint update. Blue arrows denote reads; dashed orange arrows denote writes.}
  \label{fig:state_comparison}
\end{figure}

One fast MLP contains $262{,}912$ state values. \method{} uses one such state per trajectory, whereas the layer-local control uses eight, totaling $2{,}103{,}296$ values. Figure~\ref{fig:state_comparison} contrasts these two ways of organizing the state. The comparison fixes the individual MLP architecture and update timing while allowing each local module to learn its own adaptation parameters.

\subsection{Task Groups and Averaging}
\label{app:taskgroups}

Table~\ref{tab:robotwin_groups} defines ten task groups for the state-control experiment, based on RoboTwin instructions and manipulation sequences. Each of the 50 tasks belongs to one group according to its defining interaction. For compound tasks, the distinguishing requirement takes precedence. Examples include drawer opening in \path{put_object_cabinet}, orientation in \path{place_dual_shoes}, and arm-to-arm transfer in \path{handover_block}.

Let $g(a)$ be the group of target task $a$. Similar selects uniformly among tasks $b\ne a$ with $g(b)=g(a)$; Other selects uniformly among tasks $c$ with $g(c)\ne g(a)$. Another episode of $a$ is excluded from both source pools. For example, \path{adjust_bottle} and \path{pick_diverse_bottles} share G1, whereas \path{stack_blocks_three} belongs to G5. Group membership defines the comparison; whether the imported experience helps is measured by target success.

For condition $s$, let $\mathrm{SR}(a;s)$ be the percentage of target episodes that satisfy the benchmark's success criterion. Table~\ref{tab:intervention} reports Clean$\to$Clean results only:
\begin{equation}
  \overline{\mathrm{SR}}_{s}
  = \frac{1}{50}\sum_{a\in\mathcal{T}}\mathrm{SR}(a;s),
  \qquad |\mathcal{T}|=50
  \label{eq:task_average_sr}
\end{equation}
Every target task receives equal weight; groups determine eligible history sources only. NCS is first averaged over complete target episodes within each task, then equally over tasks. Source episodes do not contribute to the target score.

\subsection{Closed-Loop State Controls}
\label{app:intervention}

Let $W_B^{\mathrm{end}}$ and $W_C^{\mathrm{end}}$ denote states accumulated over complete similar-task and other-task source episodes. For target episode $A$, the four initializations are
\begin{equation}
  W_{\mathrm{init}}^s =
  \begin{cases}
    W_0, & s\in\{\mathrm{Full},\mathrm{Reset}\},\\
    W_B^{\mathrm{end}}, & s=\mathrm{Similar},\\
    W_C^{\mathrm{end}}, & s=\mathrm{Other}
  \end{cases}
  \label{eq:state_initialization}
\end{equation}
These states initialize the target's first call. Full, Similar, and Other then apply the ordinary read--generate--update sequence; Reset restores only the fast state to $W_0$ before every call. All conditions use the same checkpoint, observation interface, action chunking, and slow parameters.

Conditions share the target task, initial environment seed, and policy sampling randomness. Each rollout runs to the benchmark's terminal success criterion or episode limit. Success is measured over the whole target episode.

Source tasks follow the pools in Appendix~\ref{app:taskgroups}, with episode seeds fixed before target evaluation. Source and target use the same checkpoint and Clean setting. Each source runs to success or its episode limit, and its final state is retained regardless of outcome. A complete history can contain different numbers of policy calls across tasks. Similar and Other each receive one prior rollout; their comparison tests its relevance to the target task.

\subsection{State-Control Metrics}
\label{app:state_metrics}

\paragraph{Success and completion steps.}
SR uses the benchmark's binary success criterion over the complete target episode. For episode $i$ of task $a$, let $T_i^s$ be the first successful environment action step under condition $s$, and let $T_a^{\max}$ be the task's evaluation step limit. Normalized completion steps are
\begin{equation}
  \mathrm{NCS}_i^s =
  \begin{cases}
    T_i^s / T_a^{\max}, & \text{if the target episode succeeds},\\
    1, & \text{otherwise}
  \end{cases}
  \label{eq:normalized_completion_steps}
\end{equation}
Lower NCS reflects a higher success rate, a smaller completion-budget fraction among successful trials, or both. Failures receive a score of one. The metric measures target execution and excludes the cost of generating source histories.

If $p$ is the success fraction and $r$ the mean $T_i^s/T_a^{\max}$ among successful trials, then $\mathrm{NCS}=1-p+pr$ under the equal-count protocol. The Table~\ref{tab:intervention} values imply $r\approx30.16\%$, $30.18\%$, $29.21\%$, and $47.53\%$ for Full, Reset, Similar, and Other, respectively. These are derived summaries of each condition's successful trials, not additional measurements.

% Keep the task inventory after its definitions; allow the next section to
% use the remaining space below the table.
% Task membership matches ../robotwin_task_groups.csv.
% Groups define eligible history sources; they are not result-averaging units.
% Permit page breaks between groups instead of leaving a half-empty page.
\begingroup
\small
\setlength{\tabcolsep}{4pt}
\renewcommand{\arraystretch}{1.16}
\setlength{\LTcapwidth}{\linewidth}
\setlength{\LTpre}{8pt}
\setlength{\LTpost}{8pt}
\begin{longtable}{@{}>{\raggedright\arraybackslash}p{0.22\linewidth}>{\raggedright\arraybackslash}p{0.24\linewidth}>{\raggedright\arraybackslash}p{\dimexpr0.54\linewidth-4\tabcolsep\relax}@{}}
\caption{\textbf{Task groups for selecting prior experience.} We assign each of the 50 RoboTwin tasks to one group by its defining manipulation requirement. Counts are task counts; identifiers follow the benchmark.}
\label{tab:robotwin_groups}\\
\toprule
Group (tasks) & Defining requirement & Task identifiers\\
\midrule
\endfirsthead
\multicolumn{3}{@{}l}{\tablename~\thetable{} (continued)}\\
\toprule
Group (tasks) & Defining requirement & Task identifiers\\
\midrule
\endhead
\midrule
\endfoot
\bottomrule
\endlastfoot
G1: Grasping and lifting (5) & Acquire and hold objects above the table &
\path{adjust_bottle}, \path{grab_roller}, \path{lift_pot}, \path{pick_diverse_bottles}, \path{pick_dual_bottles}\\
\addlinespace[3pt]
G2: Spatial placement (11) & Relocate to a surface, region, or relative position &
\path{move_can_pot}, \path{move_pillbottle_pad}, \path{move_playingcard_away}, \path{move_stapler_pad}, \path{place_a2b_left}, \path{place_a2b_right}, \path{place_container_plate}, \path{place_empty_cup}, \path{place_mouse_pad}, \path{place_object_scale}, \path{place_object_stand}\\
\addlinespace[3pt]
G3: Container loading (7) & Load items into a receptacle or onto a tray &
\path{place_bread_basket}, \path{place_bread_skillet}, \path{place_burger_fries}, \path{place_can_basket}, \path{place_cans_plasticbox}, \path{place_object_basket}, \path{put_bottles_dustbin}\\
\addlinespace[3pt]
G4: Oriented placement and mounting (6) & Set an explicitly required facing direction or engage a fixture &
\path{hanging_mug}, \path{place_dual_shoes}, \path{place_fan}, \path{place_phone_stand}, \path{place_shoe}, \path{rotate_qrcode}\\
\addlinespace[3pt]
G5: Ordering and stacking (6) & Construct an ordered arrangement or vertical stack &
\path{blocks_ranking_rgb}, \path{blocks_ranking_size}, \path{stack_blocks_three}, \path{stack_blocks_two}, \path{stack_bowls_three}, \path{stack_bowls_two}\\
\addlinespace[3pt]
G6: Arm-to-arm handover (2) & Transfer a grasped object between arms &
\path{handover_block}, \path{handover_mic}\\
\addlinespace[3pt]
G7: Pouring and shaking (3) & Manipulate a held object through repeated pouring or shaking motions &
\path{dump_bin_bigbin}, \path{shake_bottle}, \path{shake_bottle_horizontally}\\
\addlinespace[3pt]
G8: Articulated opening (3) & Open a lid, door, or drawer to complete the task &
\path{open_laptop}, \path{open_microwave}, \path{put_object_cabinet}\\
\addlinespace[3pt]
G9: Direct pressing (4) & Actuate a target directly with the gripper &
\path{click_alarmclock}, \path{click_bell}, \path{press_stapler}, \path{turn_switch}\\
\addlinespace[3pt]
G10: Tool use (3) & Act on a target through a grasped tool &
\path{beat_block_hammer}, \path{scan_object}, \path{stamp_seal}\\
\end{longtable}
\endgroup

% Keep sequence pages compact without stretching gaps around fixed figures.
\raggedbottom
\section{Real-Robot Setup and Execution Sequences}
\label{app:real}
\label{app:qualitative}

\paragraph{Platform and recordings.}
The platform uses two SOARM101 arms over a shared tabletop, with an LRCP G720P camera on each wrist. Figure~\ref{fig:real_overview} shows the setup and twelve task scenes. The sequences below are additional illustrative \method{} executions, separate from the scored trials.

\paragraph{Fine-tuning and evaluation tasks.}
All three policies use the same $550$ demonstrations from $55$ training tasks, with $10$ demonstrations per task (Table~\ref{tab:real_training}). The training set covers all nine evaluation task families. T1--T12 are held-out variants within these families, testing new operation sequences or combinations and new scene or object configurations relative to the fine-tuning set.

\begin{table}[!htbp]
\centering
\caption{\textbf{Real-robot fine-tuning data.} Each training task has ten demonstrations. The evaluation tasks are held-out variants within the same task families.}
\label{tab:real_training}
\small
\setlength{\tabcolsep}{5pt}
\renewcommand{\arraystretch}{1.12}
\begin{tabularx}{\linewidth}{@{}Xrrl@{}}
\toprule
Task family & Training tasks & Demonstrations & Evaluation tasks\\
\midrule
Button pressing & 10 & 100 & T1--T3\\
Fruit packing & 10 & 100 & T4\\
Block stacking & 5 & 50 & T5--T6\\
Ball placement & 5 & 50 & T8\\
Paper-cup stacking & 5 & 50 & T9\\
Towel folding & 5 & 50 & T10\\
Cup manipulation & 5 & 50 & T11\\
Toothbrush and toothpaste placement & 5 & 50 & T7\\
Table wiping & 5 & 50 & T12\\
\midrule
Total & 55 & 550 & T1--T12\\
\bottomrule
\end{tabularx}
\end{table}

\paragraph{Success criteria.}
For T1--T3, success requires the exact instructed sequence and number of presses, followed by $10$\,s without an additional press while the policy remains active. Stacking succeeds when the cubes form the required stack, with T5 additionally requiring the specified color grouping. T10 is judged by the final towel configuration folded into thirds, which can be formed by two successive folds. T12 requires sponge contact with the tabletop during the prescribed left/right and then up/down wiping motions. The T4 fruit and box perturbations appear only in the illustrative executions and are excluded from the quantitative trials.

\paragraph{Complete task results.}
Table~\ref{tab:real_full} reports success counts for all twelve tasks, with $20$ trials per policy and task. Its totals determine the averages in Table~\ref{tab:real}; the execution sequences below illustrate behaviors and failures separately from this quantitative summary.

\begin{table}[H]
\centering
\caption{\textbf{Complete real-robot success counts.} Entries are successes out of 20 trials. All twelve tasks contribute equally to the average.}
\label{tab:real_full}
\small
\setlength{\tabcolsep}{5pt}
\renewcommand{\arraystretch}{1.15}
\begin{tabularx}{\linewidth}{@{}lXrrr@{}}
\toprule
Task & Instruction & $\pi_{0.5}$ & Matched Base & \method{}\\
\midrule
T1 & Orange once, then blue twice & 6/20 & 4/20 & 15/20\\
T2 & Blue once, then orange twice & 5/20 & 5/20 & 13/20\\
T3 & Press blue three times & 3/20 & 1/20 & 12/20\\
T4 & Pack fruit, banana last & 9/20 & 7/20 & 12/20\\
T5 & Stack blocks by color & 10/20 & 9/20 & 12/20\\
T6 & Stack blocks & 9/20 & 6/20 & 10/20\\
T7 & Place toothbrush, then toothpaste & 9/20 & 4/20 & 7/20\\
T8 & Place balls into a paper cup & 12/20 & 10/20 & 12/20\\
T9 & Stack paper cups in order & 9/20 & 6/20 & 11/20\\
T10 & Fold a towel into thirds & 6/20 & 4/20 & 8/20\\
T11 & Orient a cup and remove its lid & 11/20 & 8/20 & 7/20\\
T12 & Wipe the table with a sponge & 13/20 & 10/20 & 12/20\\
\midrule
 & Total & 102/240 & 74/240 & 131/240\\
 & Mean success (\%) & 42.50 & 30.83 & 54.58\\
\bottomrule
\end{tabularx}
\end{table}

\begin{figure}[H]
  \centering
  \includegraphics[width=\linewidth]{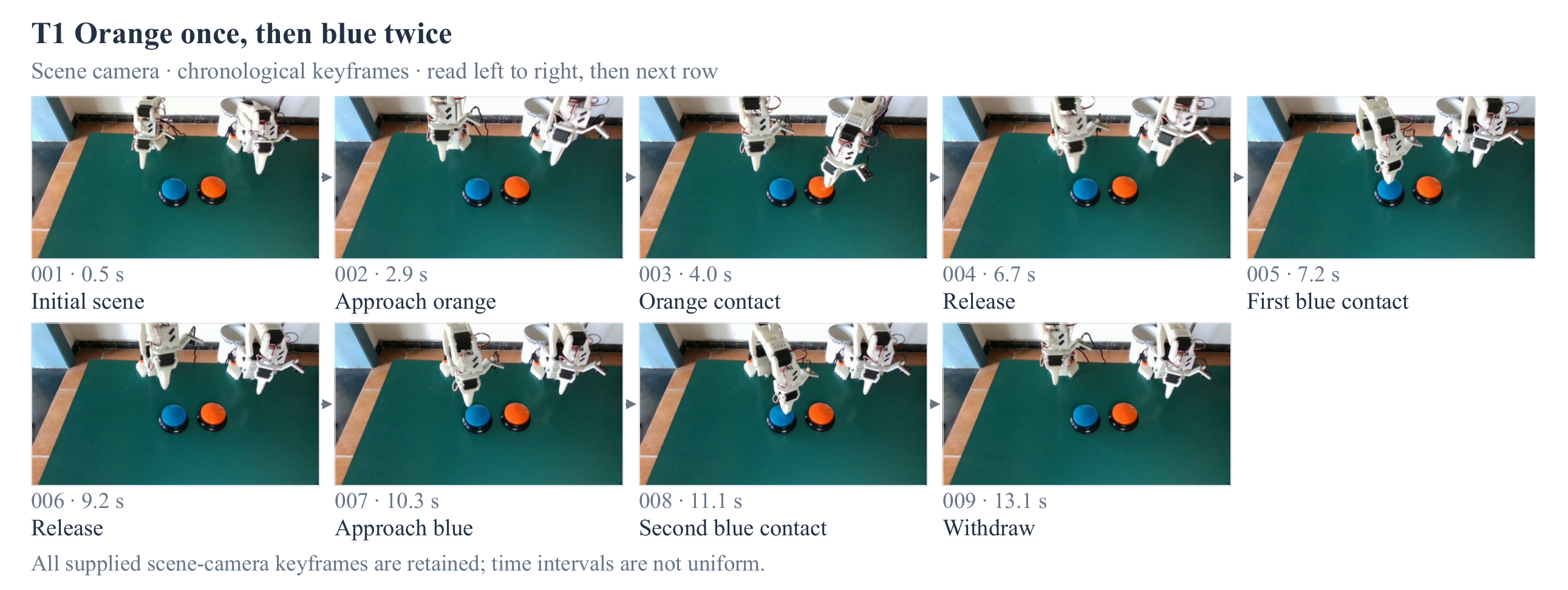}
  \caption{\textbf{T1: Orange once, then blue twice.} Orange contact precedes two blue-contact phases separated by withdrawal.}
  \label{fig:real_t1}
\end{figure}

\begin{figure}[H]
  \centering
  \includegraphics[width=\linewidth]{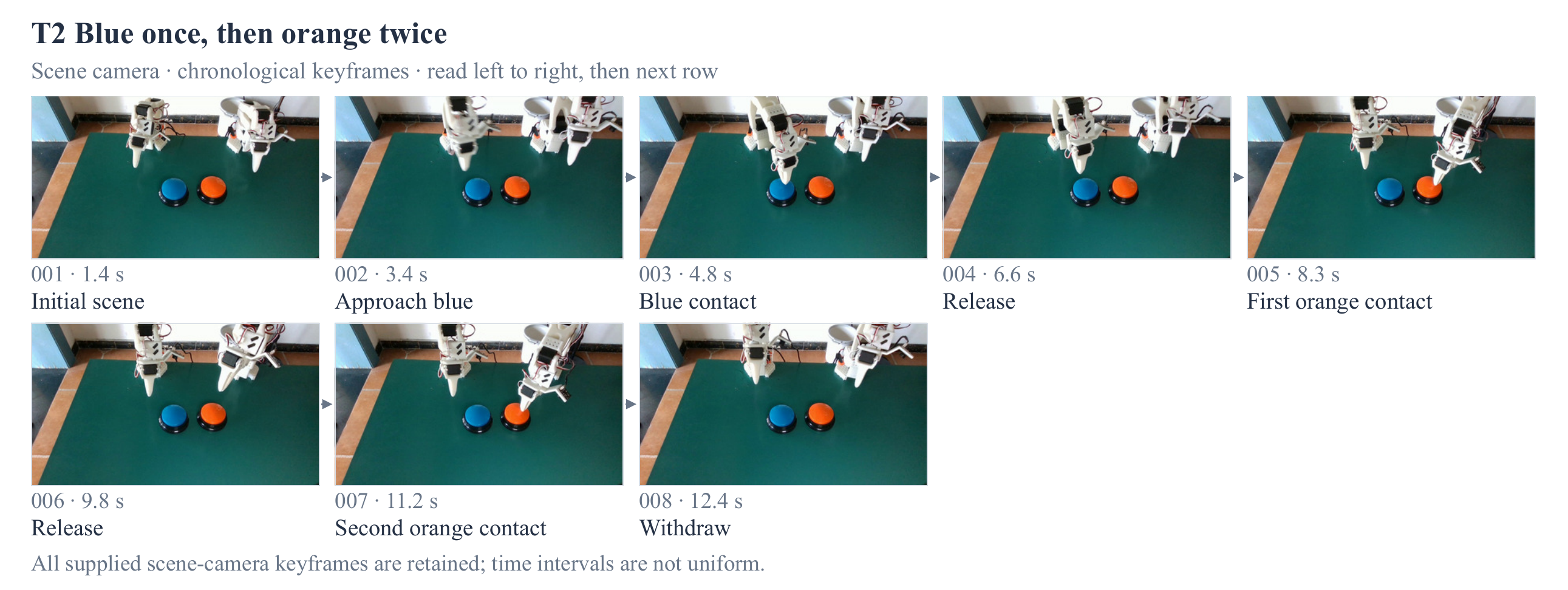}
  \caption{\textbf{T2: Blue once, then orange twice.} The robot switches from blue to orange and revisits orange after withdrawal.}
  \label{fig:real_t2}
\end{figure}

\begin{figure}[H]
  \centering
  \includegraphics[width=\linewidth]{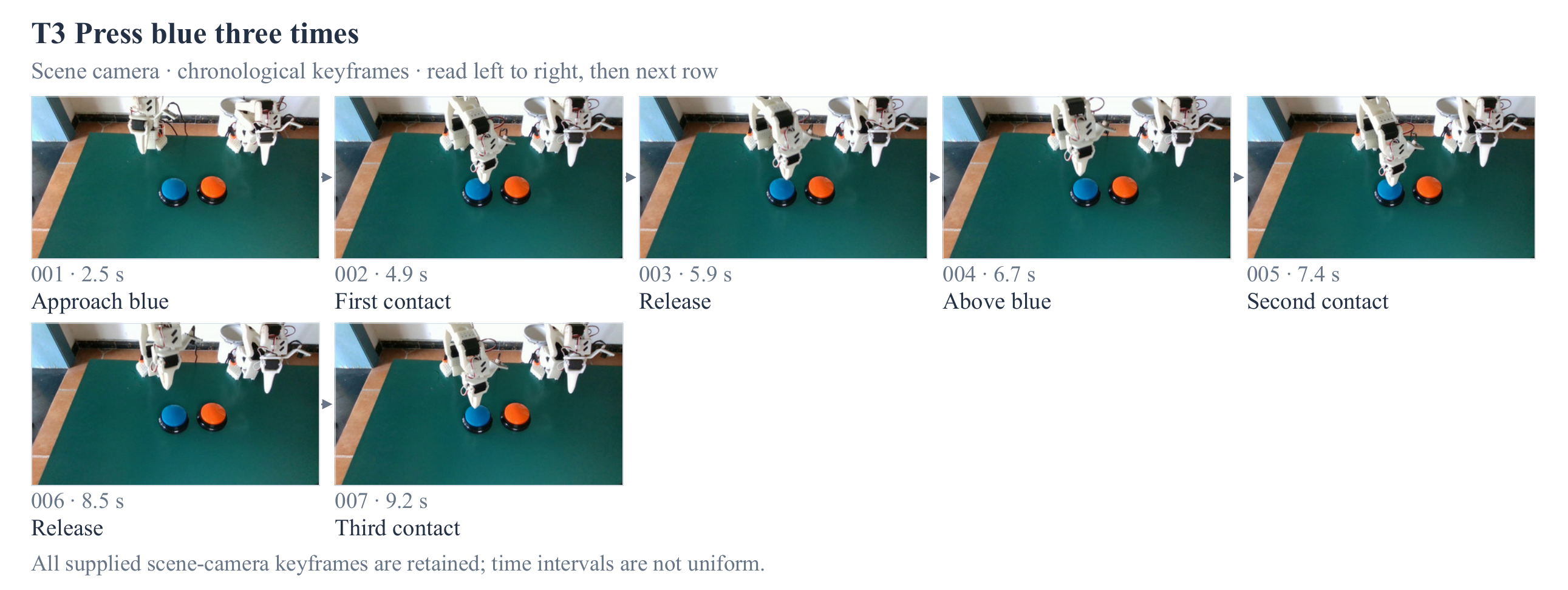}
  \caption{\textbf{T3: Press blue three times.} Repeated contact and withdrawal at the same button.}
  \label{fig:real_t3}
\end{figure}

\begin{figure}[H]
  \centering
  \includegraphics[width=\linewidth]{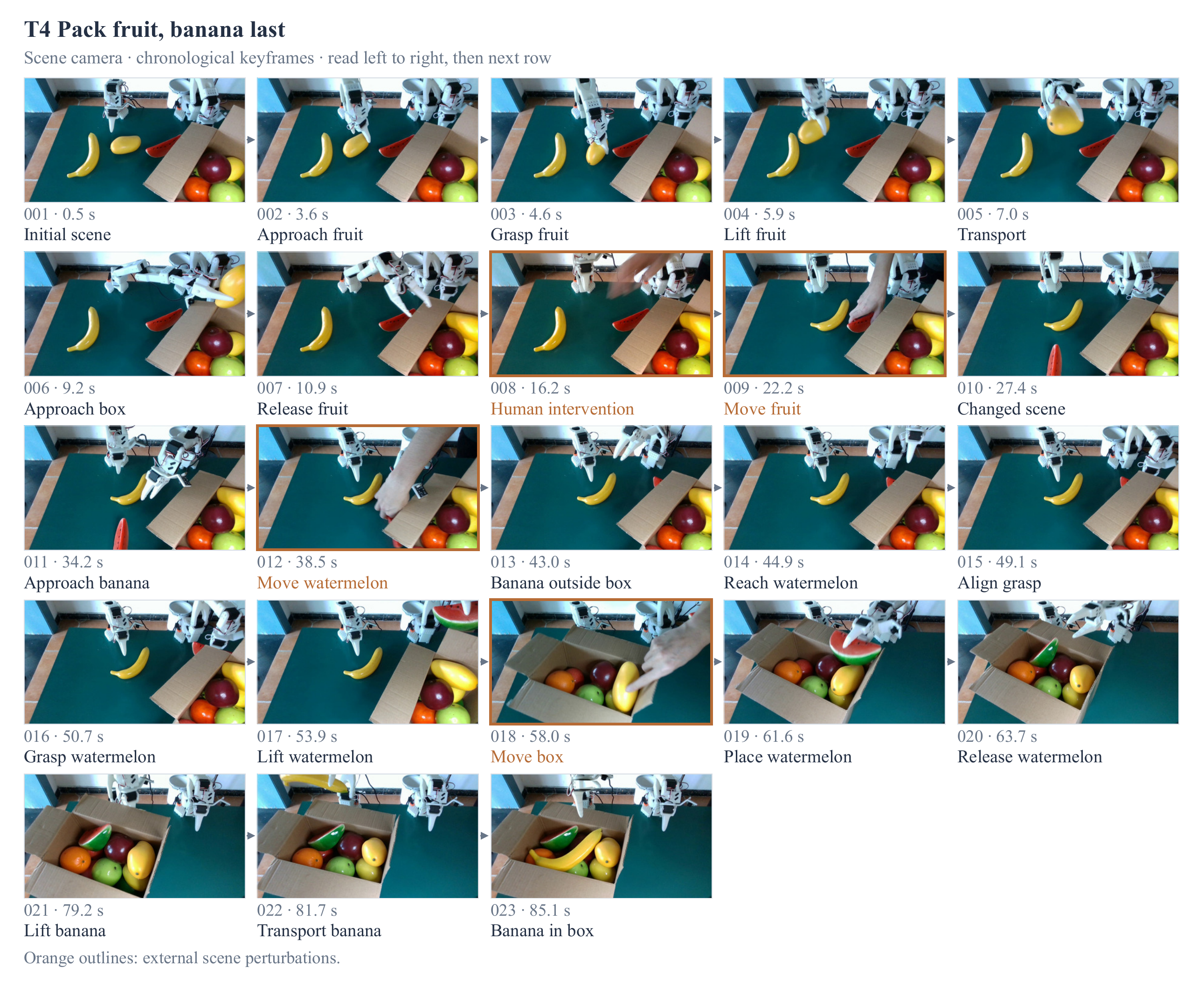}
  \caption{\textbf{T4: Pack fruit, banana last.} The robot preserves the ordering constraint after human interventions (orange outlines).}
  \label{fig:real_t4}
\end{figure}

\begin{figure}[H]
  \centering
  \includegraphics[width=\linewidth]{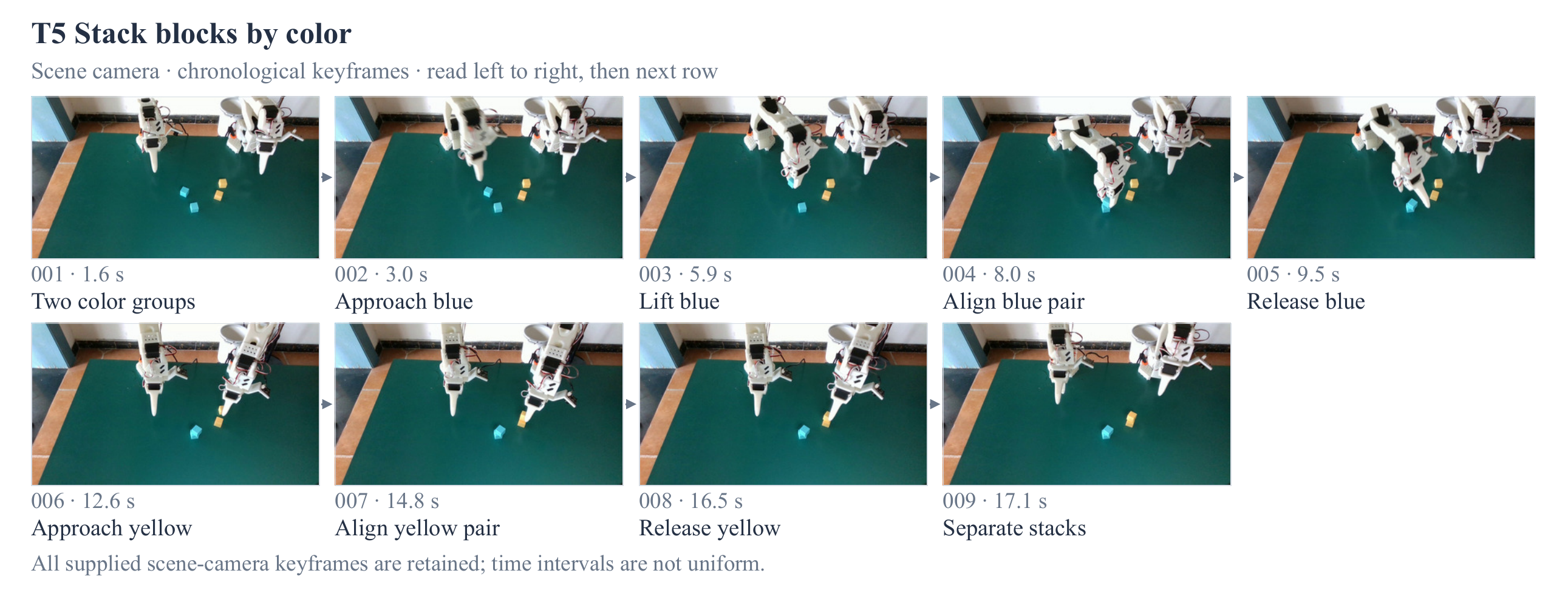}
  \caption{\textbf{T5: Color-conditioned stacking.} Blue and yellow pairs are assembled into separate stacks.}
  \label{fig:real_t5}
\end{figure}

\begin{figure}[H]
  \centering
  \includegraphics[width=\linewidth]{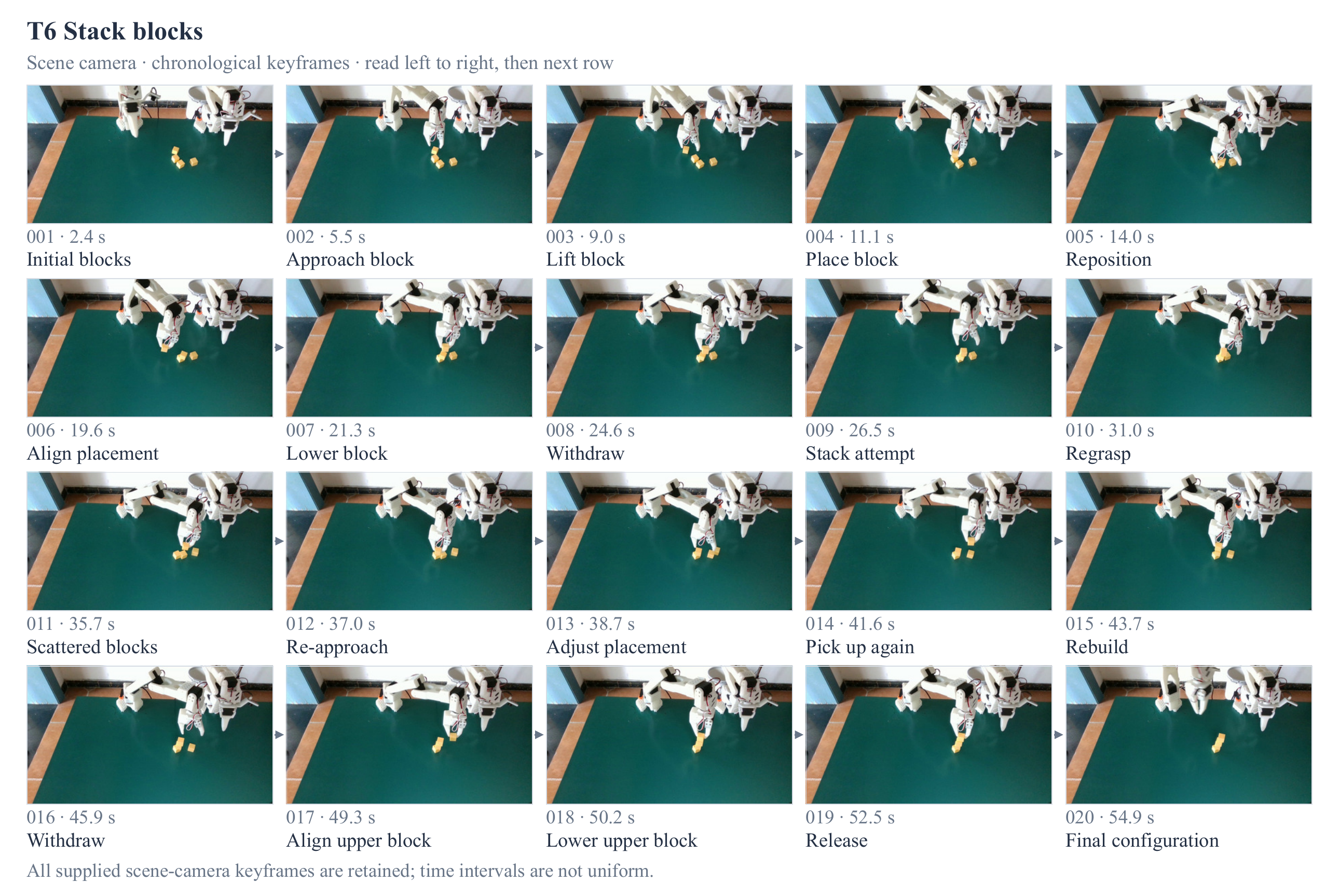}
  \caption{\textbf{T6: Repeated stacking attempts and recovery.} The robot resumes placement after the intermediate stack collapses.}
  \label{fig:real_t6}
\end{figure}

\begin{figure}[H]
  \centering
  \includegraphics[width=\linewidth]{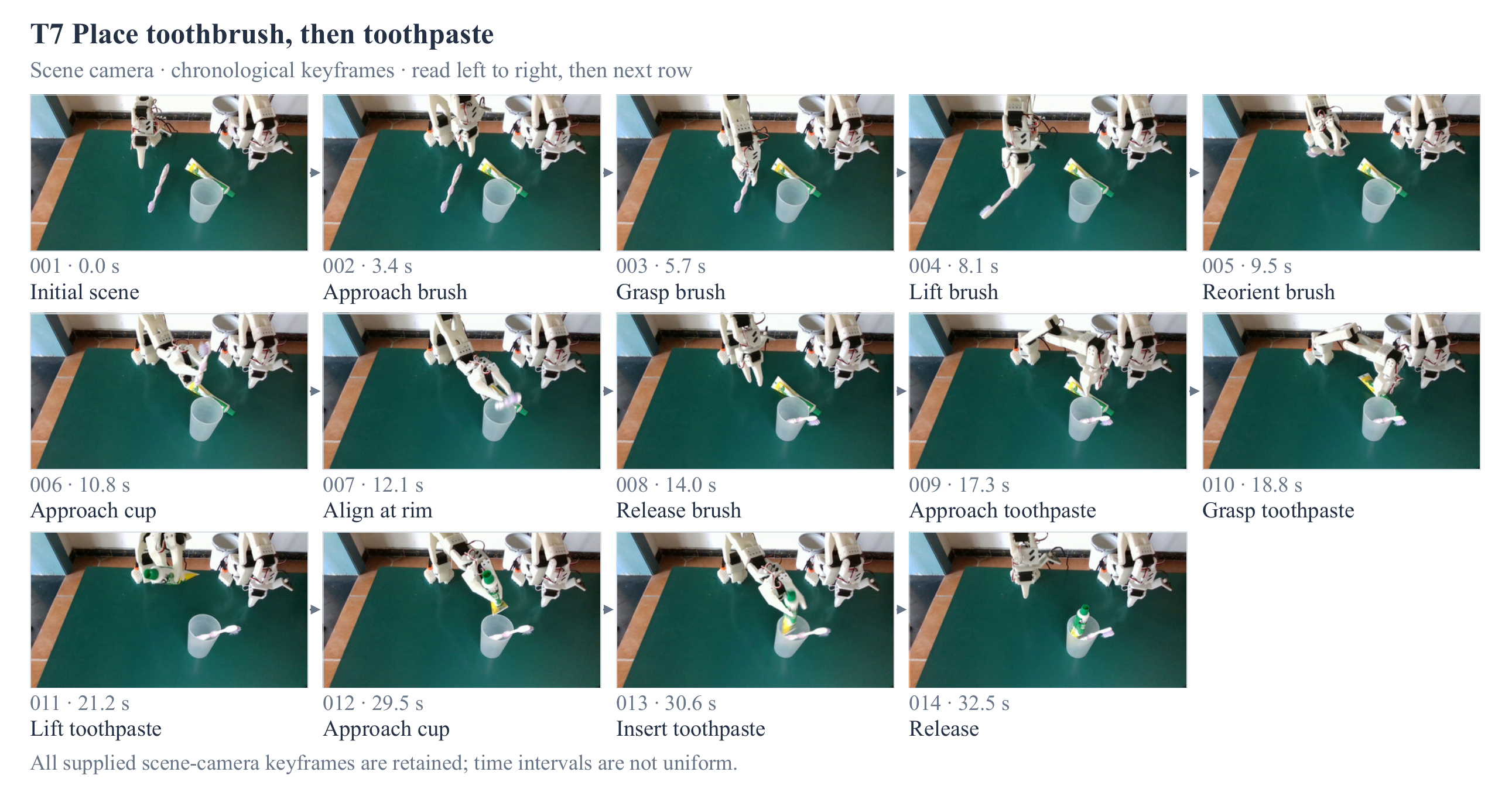}
  \caption{\textbf{T7: Object order and insertion geometry.} Toothbrush handling precedes toothpaste handling, but the released toothbrush lies across the cup rim.}
  \label{fig:real_t7}
\end{figure}

\begin{figure}[H]
  \centering
  \includegraphics[width=\linewidth]{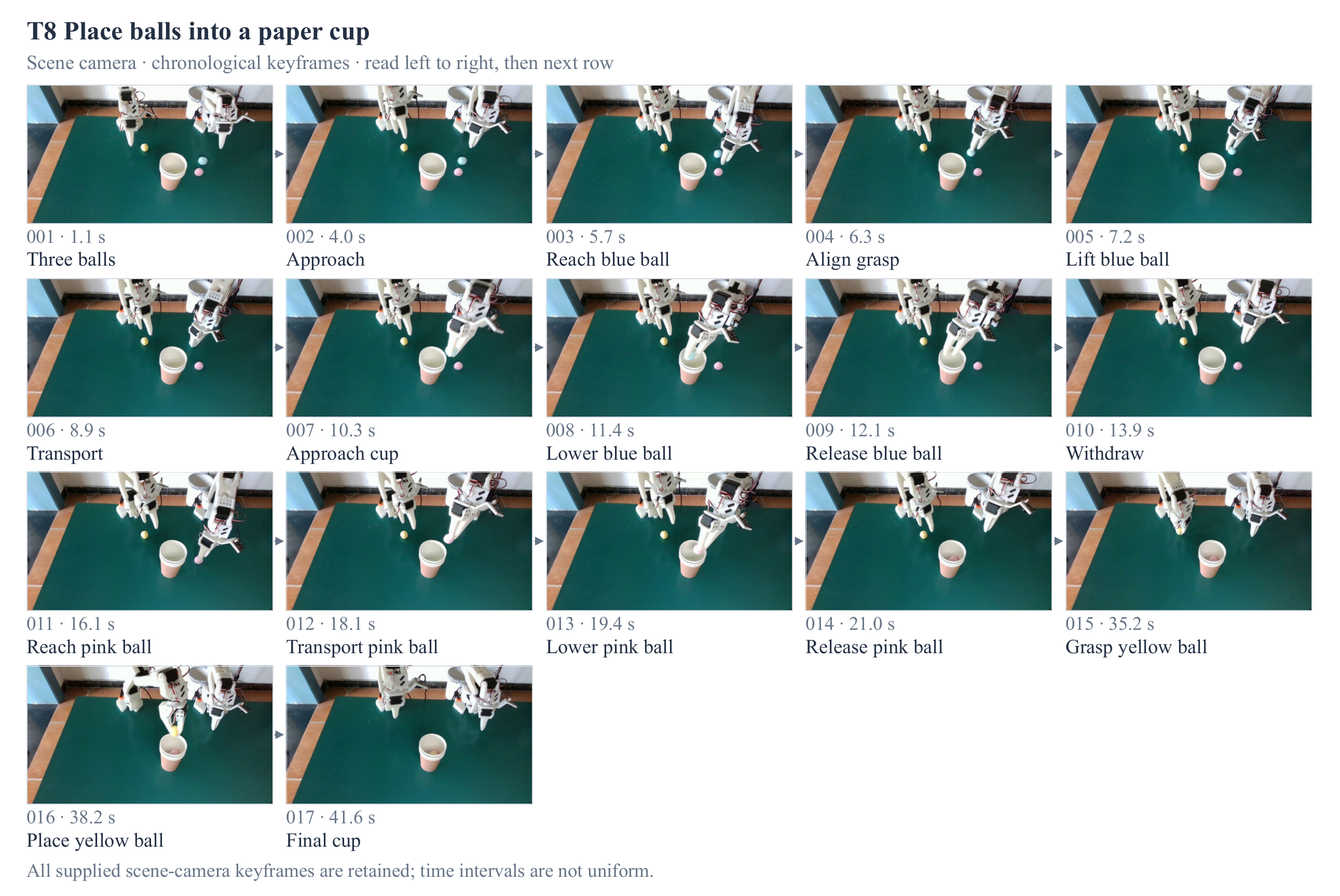}
  \caption{\textbf{T8: Place balls into a paper cup.} Three balls are successively grasped and placed into the cup.}
  \label{fig:real_t8}
\end{figure}

\begin{figure}[H]
  \centering
  \includegraphics[width=\linewidth]{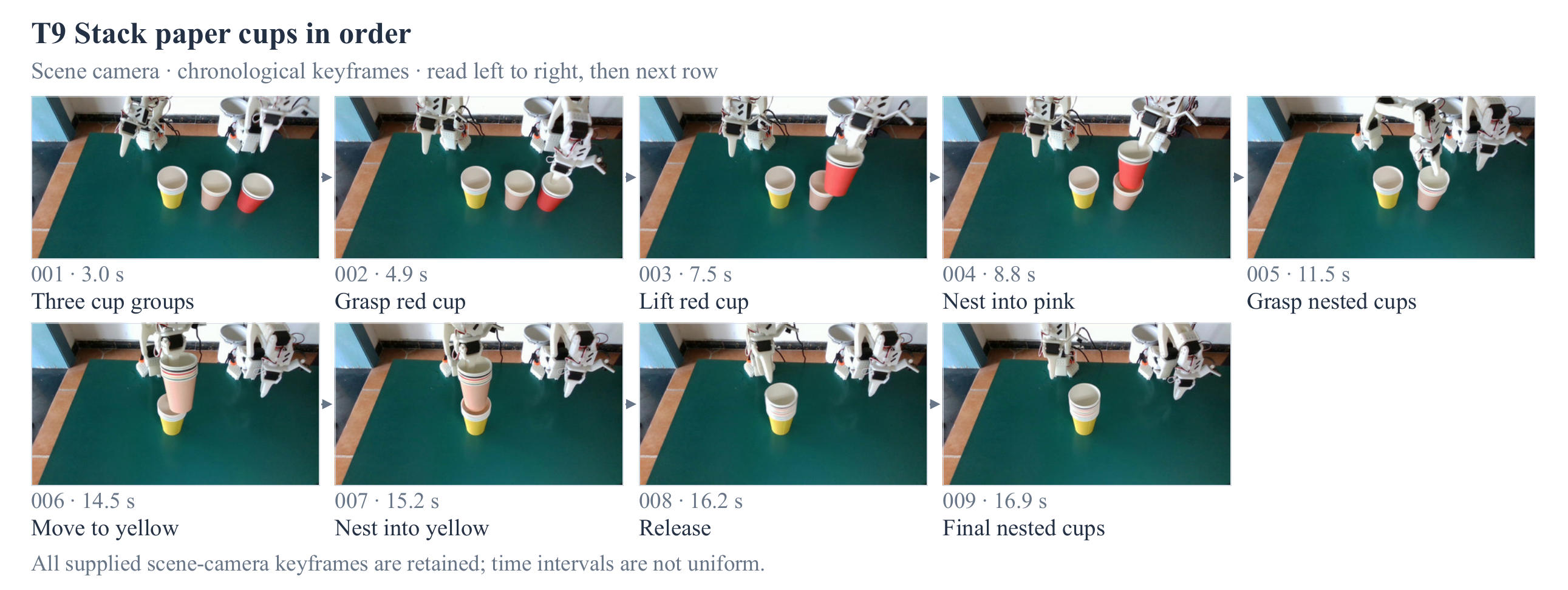}
  \caption{\textbf{T9: Successive cup nesting.} Red is nested into pink, and the combined assembly is then nested into yellow.}
  \label{fig:real_t9}
\end{figure}

\begin{figure}[H]
  \centering
  \includegraphics[width=\linewidth]{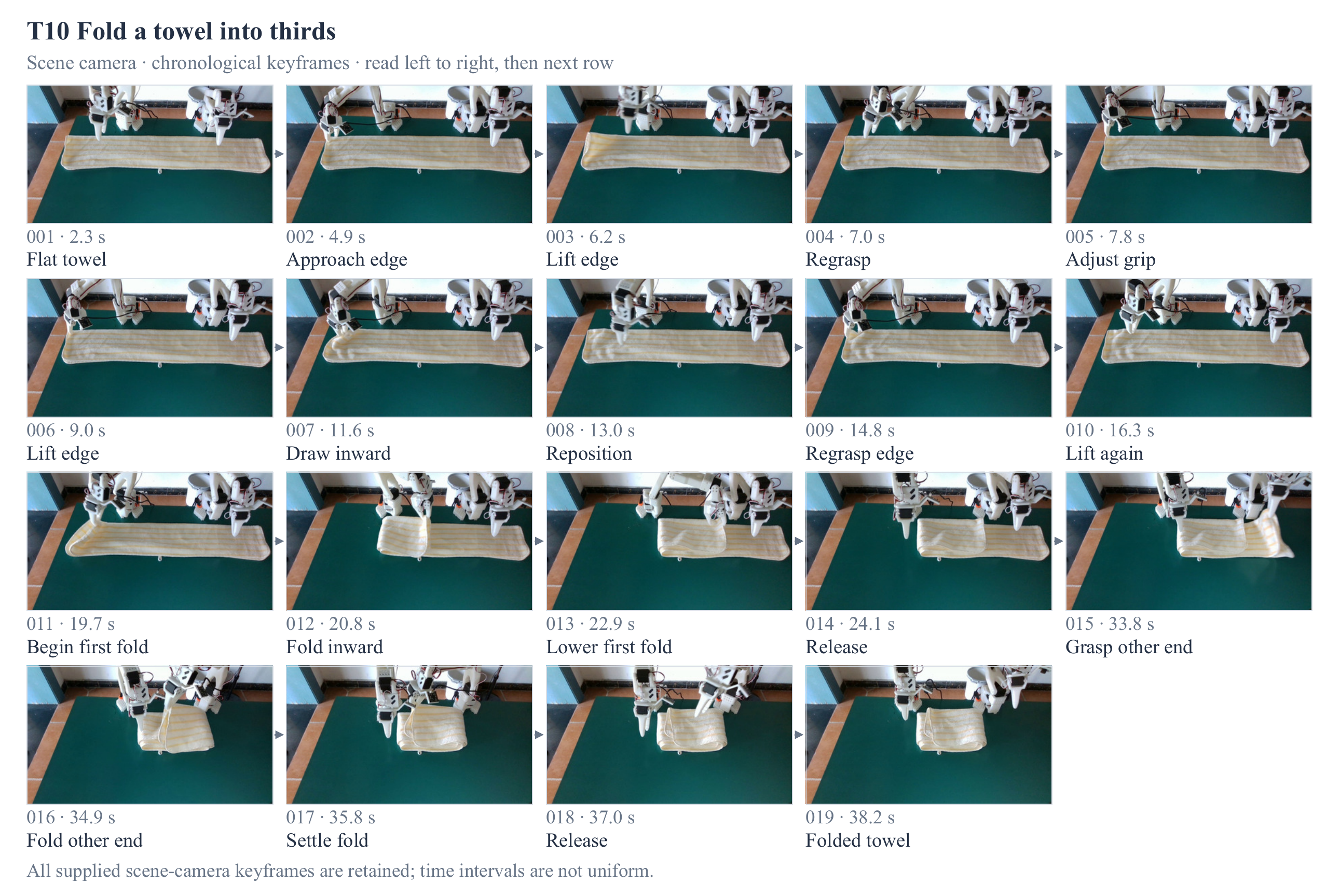}
  \caption{\textbf{T10: Two successive folds with grasp adjustments.} Repeated edge-grasp attempts precede folding from opposite ends.}
  \label{fig:real_t10}
\end{figure}

\begin{figure}[H]
  \centering
  \includegraphics[width=\linewidth]{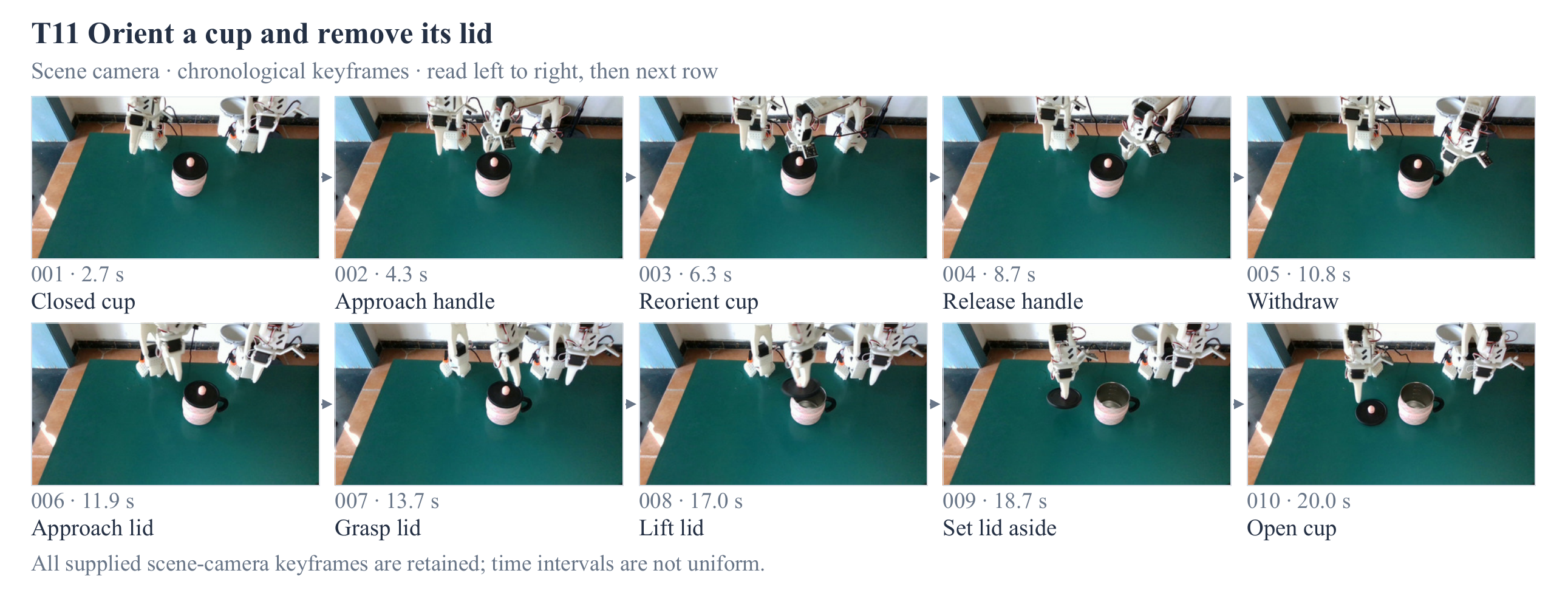}
  \caption{\textbf{T11: Orient a cup and remove its lid.} Handle-guided reorientation precedes lid removal and placement beside the cup.}
  \label{fig:real_t11}
\end{figure}

\begin{figure}[H]
  \centering
  \includegraphics[width=\linewidth]{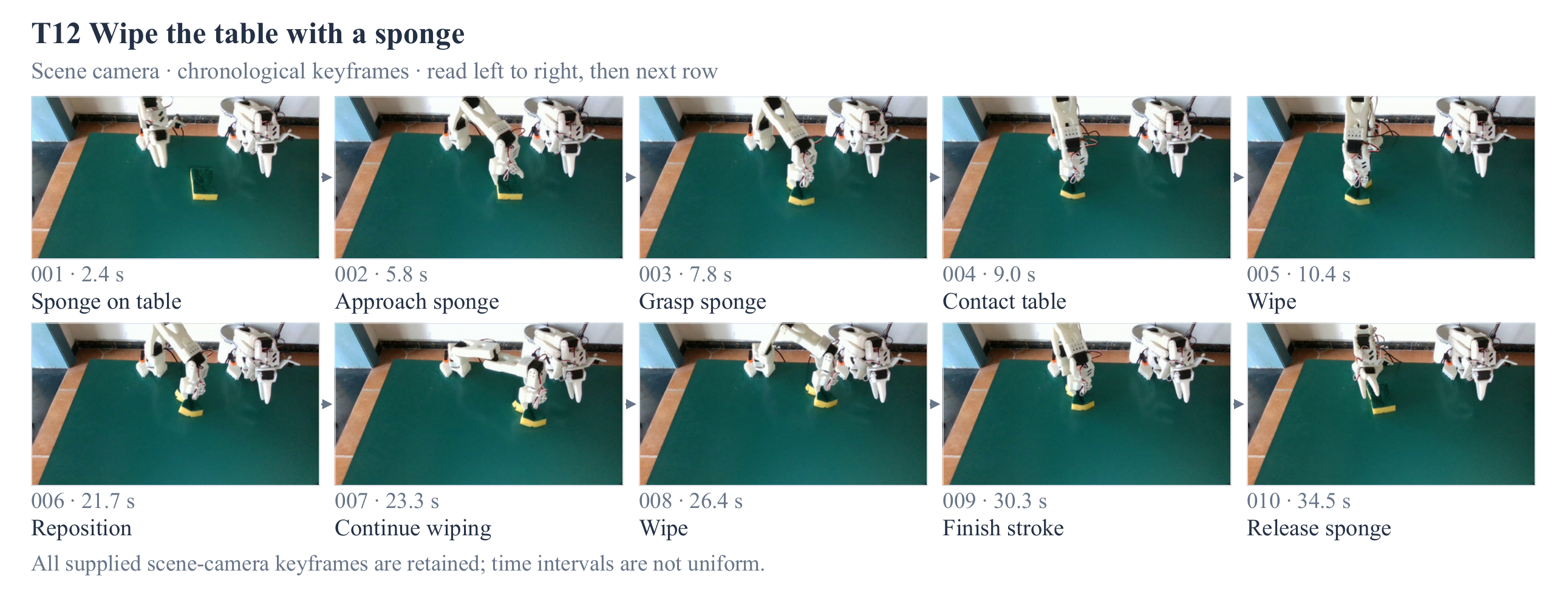}
  \caption{\textbf{T12: Sponge grasping and tabletop wiping.} The views show the approach, successive wiping positions, and release.}
  \label{fig:real_t12}
\end{figure}

\paragraph{Reading the sequences.}
Figures~\ref{fig:real_t1}--\ref{fig:real_t12} show scene-camera keyframes in temporal order, with five images per row. Read left to right, then continue on the next row. Labels give keyframe IDs and elapsed times; intervals vary across frames. Orange outlines in T4 mark external perturbations. Task descriptions distinguish visible events from additional failures and interventions reported in the execution notes.

\paragraph{History and perception.}
After a button is released, the scene can resemble an earlier step, whereas packing and stacking often leave visible evidence of progress. These tasks thus place different demands on history and current perception. Fruit packing probes an ordering constraint under scene changes; stacking distinguishes recovery after collapse from accurate placement and physical stability.

% Keep each heading with its description; the sequence may continue on the next page.
\begin{samepage}
\subsection{T1: Orange Once, Then Blue Twice}
\label{app:real_t1}

The instruction requires the sequence orange--blue--blue. Figure~\ref{fig:real_t1} shows orange contact at 4.0\,s, followed by blue-contact phases at 7.2 and 11.1\,s, separated by withdrawal. Additional executions include a return to orange after the initial press, violating the instructed sequence.

\par\end{samepage}

% Keep each heading with its description; the sequence may continue on the next page.
\begin{samepage}
\subsection{T2: Blue Once, Then Orange Twice}
\label{app:real_t2}

Reversing the first color changes the required continuation to blue--orange--orange. Figure~\ref{fig:real_t2} shows blue contact at 4.8\,s and orange-contact phases at 8.3 and 11.2\,s. A reported failure is returning to blue after the initial press. Together, T1 and T2 separate button identity from instruction progress: the same objects require different continuations.

\par\end{samepage}

% Keep each heading with its description; the sequence may continue on the next page.
\begin{samepage}
\subsection{T3: Press Blue Three Times}
\label{app:real_t3}

This task fixes the target identity and requires tracking the number of completed presses. Contact phases appear at 4.9, 7.4, and 9.2\,s in Figure~\ref{fig:real_t3}. The execution notes report failures to track this count. Completion requires both returning to the same button and stopping after the requested number of presses.

\par\end{samepage}

% Keep each heading with its description; the sequence may continue on the next page.
\begin{samepage}
\subsection{T4: Pack Fruit, Banana Last}
\label{app:real_t4}

The task requires placing fruit in a box while reserving the banana for the end. In Figure~\ref{fig:real_t4}, fruit is repositioned at 22.2\,s, the watermelon is moved at 38.5\,s, and the box shifts at 58.0\,s. The robot places the watermelon at 61.6\,s and the banana at 85.1\,s, preserving the ordering constraint despite these scene changes.

The execution notes additionally describe the robot putting down a substituted banana and resuming the interrupted fruit placement, as well as retrieving fruit after partial concealment. These behaviors illustrate maintaining the required order as object locations change.

\par\end{samepage}

% Keep each heading with its description; the sequence may continue on the next page.
\begin{samepage}
\subsection{T5: Stack Blocks by Color}
\label{app:real_t5}

Two blue blocks and two yellow blocks must form separate same-color stacks. Figure~\ref{fig:real_t5} first assembles the blue pair and then the yellow pair. The execution notes report color mixing as a characteristic failure, violating the grouping constraint even when grasping and placement are feasible.

\par\end{samepage}

% Keep each heading with its description; the sequence may continue on the next page.
\begin{samepage}
\subsection{T6: Stack Blocks}
\label{app:real_t6}

The robot makes repeated attempts to build a taller stack. The arrangement collapses around 31.0--38.7\,s, followed by renewed grasping and placement (Figure~\ref{fig:real_t6}). The execution notes identify instability and arm contact as characteristic causes. The robot can resume stacking after a collapse, but alignment, clearance, and physical stability remain limiting factors.

\par\end{samepage}

% Keep each heading with its description; the sequence may continue on the next page.
\begin{samepage}
\subsection{T7: Place Toothbrush, Then Toothpaste}
\label{app:real_t7}

The robot handles the toothbrush first and the toothpaste second (Figure~\ref{fig:real_t7}). After release, the toothbrush rests across the cup rim; the toothpaste is then lowered into the opening. The robot follows the required order but fails to place the toothbrush securely, illustrating the difficulty of aligning objects with the cup opening.

\par\end{samepage}

% Keep each heading with its description; the sequence may continue on the next page.
\begin{samepage}
\subsection{T8: Place Balls into a Paper Cup}
\label{app:real_t8}

Blue, pink, and yellow balls are handled in succession and lowered into a paper cup (Figure~\ref{fig:real_t8}). The task combines small-object grasping with repeated alignment to one receptacle. The changing set of tabletop objects can provide a visible progress cue for subsequent placements.

\par\end{samepage}

% Keep each heading with its description; the sequence may continue on the next page.
\begin{samepage}
\subsection{T9: Stack Paper Cups in Order}
\label{app:real_t9}

The robot nests the red cup into the pink group, then places the combined assembly into the yellow group (Figure~\ref{fig:real_t9}). The first nesting operation changes the grasped object. The second therefore requires transporting and aligning a larger combined assembly.

\par\end{samepage}

% Keep each heading with its description; the sequence may continue on the next page.
\begin{samepage}
\subsection{T10: Fold a Towel into Thirds}
\label{app:real_t10}

The robot adjusts its grasp on one towel end before bringing it inward at 19.7--24.1\,s. It then folds the opposite end over the first fold at 33.8--38.2\,s (Figure~\ref{fig:real_t10}). The sequence includes local retries and a transition between two manipulation phases on a deformable object.

\par\end{samepage}

% Keep each heading with its description; the sequence may continue on the next page.
\begin{samepage}
\subsection{T11: Orient a Cup and Remove Its Lid}
\label{app:real_t11}

The robot first uses the handle to rotate the cup, then grasps the lid knob, lifts the lid, and sets it beside the cup (Figure~\ref{fig:real_t11}). The task requires switching contact targets on the same object, from the handle for reorientation to the knob for lid removal.

\par\end{samepage}

% Keep each heading with its description; the sequence may continue on the next page.
\begin{samepage}
\subsection{T12: Wipe the Table with a Sponge}
\label{app:real_t12}

The sequence shows sponge grasping, movement over the tabletop, and release (Figure~\ref{fig:real_t12}). The robot carries the sponge through multiple wiping positions before releasing it, illustrating an extended manipulation phase with a grasped tool.

\par\end{samepage}

\end{document}